\documentclass[acmlarge]{acmart}
\makeatletter
\newcommand{\confshort}{\acmConference@shortname}
\newcommand{\conffull}{\acmConference@name}
\newcommand{\confdate}{\acmConference@date}
\newcommand{\confloc}{\acmConference@venue}
\AtBeginDocument{
  \fancypagestyle{firstpagestyle}{
    \fancyhead{}%
    \fancyfoot[C]{}%
  }
  \fancyhf{}
  \fancyhead[LO]{\@headfootfont\shorttitle}%
  \fancyhead[RE]{\@headfootfont\@shortauthors}%
  \fancyhead[LE]{\@headfootfont\footnotesize \confshort, \confdate, \confloc}%
  \fancyhead[RO]{\@headfootfont\footnotesize \confshort, \confdate, \confloc}%
  \fancyfoot[C]{}%
}
\makeatother
\acmBooktitle{\conffull\@ (\confshort), \confdate, \confloc}

\usepackage{booktabs}
\usepackage{longtable}
\usepackage{multirow}
\usepackage{ragged2e}
\usepackage{array}
\usepackage{xcolor}
\usepackage{placeins}

\AtBeginDocument{%
  }

\copyrightyear{2026}
\acmYear{2026}
\setcopyright{cc}
\setcctype{by}
\acmConference[FAccT '26]{The 2026 ACM Conference on Fairness, Accountability, and Transparency}{June 25--28, 2026}{Montreal, QC, Canada}
\acmBooktitle{The 2026 ACM Conference on Fairness, Accountability, and Transparency (FAccT '26), June 25--28, 2026, Montreal, QC, Canada}
\acmDOI{10.1145/3805689.3812240}
\acmISBN{979-8-4007-2596-8/2026/06}

\begin{document}

\title{Unsteady Metrics and Benchmarking Cultures of AI Model Builders}


\author{Stefan Baack}
\authornote{All authors contributed equally to this research.}
\email{hi@sbaack.com}
\orcid{0000-0002-2464-7699}
\affiliation{%
  \institution{Independent Researcher}
}

\author{Christo Buschek}
\authornotemark[1]
\authornote{The work was conducted over the course of a Mozilla Fellowship.}
\email{christo.buschek@proton.me}
\orcid{0009-0001-4096-7465}
\affiliation{%
  \institution{Independent Researcher} 
}

\author{Maty Bohacek}
\authornotemark[1]
\email{maty@stanford.edu}
\orcid{0000-0001-8683-3692}
\affiliation{%
  \institution{Stanford University}
}

\renewcommand{\shortauthors}{Baack et al.}

\begin{abstract}

The primary way to establish and compare model competencies in foundation and generative AI models has largely moved from peer-reviewed literature to press releases and company blog posts, where model builders highlight results on a selection of benchmarks. These public-facing industry artifacts now largely define the state of the art, both for the research community and the broader public. Despite their prominence, which benchmarks model builders choose to highlight and what they communicate through this selection is underexamined. To investigate this, we introduce and open-source \emph{Benchmarking-Cultures-25}, a dataset containing $231$ benchmarks highlighted across $139$ model releases in 2025 from $11$ major AI model builders. Additionally, we publish an interactive tool to visually explore the relationships of the collected data. Our analysis points to a fragmented evaluation landscape with limited cross-model comparability: $63.2\%$ of highlighted benchmarks are used by a single model builder, and $38.5\%$ appear in just one model release. Few benchmarks achieve true widespread use (e.g., GPQA Diamond, LiveCodeBench, and AIME 2025). Moreover, benchmarks are attributed different competencies by different model builders, depending on their narrative. To disentangle these conflicting presentations, we develop a unified taxonomy that maps diverging terminology to a shared framework of measured signals based on what benchmark authors claim to measure. "General knowledge application" is the second most popular, yet vaguely defined, category of benchmark in our dataset. A qualitative analysis of these benchmarks revealed that many deemphasize construct validity; instead, they frame their results as indicators of progress toward Artificial General Intelligence (AGI). This framing is evident both in benchmarks that explicitly cite AGI literature and in those implicitly shaped by its surrounding narratives. In addition, authors of "General knowledge application" benchmarks claim to measure knowledge or reasoning capabilities in general, yet mostly evaluate them across STEM subjects (especially math). Based on these findings, we argue that highlighted benchmarks in model release artifacts currently function less as standardized measurement tools and more as flexible narrative devices that are used to construct a story of progress that prioritizes market positioning over practical scientific evaluation and comparison. Data is available at \url{https://hf.co/datasets/matybohacek/benchmarking-cultures-25}; the interactive tool is available at \url{https://bench-cultures.net}.\end{abstract}




\maketitle

\section{Introduction}

Recent work has increasingly questioned whether commonly used AI model benchmarks meaningfully reflect real-world model performance and user experience~\cite{alzahrani2024benchmarks,cheng2025survey,eriksson2025can,ethayarajh2020utility,bowman2021will,raji2021ai}. Despite these concerns, model builders continue to highlight benchmark results prominently outside academic venues---in system cards, press releases, and company blogs---for each model release~\cite{o1_system_card_2024,gpt4_research_blog_2023,claude_3_7_sonnet_system_card_2025,gpt4_system_card_2023}. The benchmarks highlighted in these public-facing industry artifacts are unlikely to reflect the full internal evaluation suite used by the respective organizations~\cite{wan20252025,bommasani20242024,haimes2024benchmark}; rather, they constitute a curated subset presented to external audiences (including prospective users and developers utilizing the models through an API), highlighting unique competencies and competitive positioning~\cite{joaquin2025deprecating}.

Although there is a substantial body of scholarship studying the quality and coverage of individual benchmarks~\cite{bean2025measuring}, as well as their usage in the academic literature~\cite{koch2021reduced,wang2024benchmark,liao2021we}, comparatively little attention has been paid to how benchmarks are selectively used by model builders to communicate model competencies in their public-facing release artifacts. Analyzing benchmarks in such contexts is an opportunity to evaluate whether they facilitate meaningful cross-model comparison and to shed light on the narratives that model builders develop through the selection of benchmarks, as this encodes implicit priorities, organizational norms, and competitive pressures.

In this paper, we construct and analyze \emph{Benchmarking-Cultures-25}, a dataset of $231$ benchmarks highlighted by $11$ prominent model builders in $139$ model releases throughout 2025. We open-source this dataset at \url{https://hf.co/datasets/matybohacek/benchmarking-cultures-25}, with an interactive web interface at \url{https://bench-cultures.net}. To construct this dataset, we devise a unified taxonomy based on what benchmark authors claim to measure to bridge the diverging terminology used by AI model builders to quantitatively analyze trends and compare how various types of model providers highlight benchmarks. Finally, we also conduct a qualitative analysis of the papers introducing the five most popular "General knowledge application" benchmarks. We address the following research questions:

\begin{itemize}
    \item[\textit{(RQ1)}] What is the makeup of benchmark author affiliations (e.g., industry, academia, government) and how is it changing over time?
    \item[\textit{(RQ2)}] Which tested competencies are the most prominent among the benchmarks, and how consistently are these competencies presented?
    \item[\textit{(RQ3)}] What are the most popular benchmarks among AI model builders?
    \item[\textit{(RQ4)}] How fast and extensively do benchmarks get adopted, and does this allow for cross-model comparison?
\end{itemize}

\section{Related Work}
\label{sec:related_work}

In addition to serving as artifacts for measuring AI model performance and progress, benchmarks also function as a technology of governance. They exert social pressure by defining hierarchies of performance, defining priorities, and ultimately compelling model builders to align with these standardized metrics (in certain cases resulting in institutional isomorphism)~\cite{wang2024benchmark,raji2021ai, dimaggio1983iron}. Due to their importance, a standalone field, often called "the science of benchmarking", has emerged, studying their mechanics, quality, and impact~\cite{laskar2024systematic,chang2024survey,liang2022holistic}. \citet{campolo2025state} situates benchmarking within a broader temporal and cultural logic, arguing that the practice of declaring state-of-the-art results functions not merely as a scientific claim but as a performative act that shapes research agendas and competitive dynamics. Relatedly, \citet{sculley2018winner} caution that the emphasis on leaderboard rankings and incremental benchmark gains risks a "winner's curse," where apparent progress on metrics obscures the absence of deeper scientific understanding. In this section, we review existing scholarship in this and adjacent fields.

\subsection{Benchmark Saturation and Goodhart's Law}

AI model builders optimize performance on benchmark metrics: in the less severe case, this occurs due to the knowledge of how testing methodologies look like, or in the more severe case through data contamination, i.e. by explicitly training on the benchmark contents (test set)~\cite{dominguez2024training,oren2023proving,ni2025training}. According to Goodhart's Law~\cite{goodhart1984problems,strathern1997improving}, such metrics cease to be informative. As a result of this direct optimization, combined with factors such as the static nature of benchmarks\footnote{Most popular benchmarks are static: they utilize a fixed, publicly-known test set that never changes after its original publication. Hybrid benchmarks, on the other hand, update their test sets over time~\cite{chen2025benchmarking}, and hence mitigate AI models' ability to learn directly on this data. This comes at the cost increased creation complexity and the need to re-run evaluations to enable back-comparability.} and slow publishing cycles\footnote{For prominent AI conferences (e.g., NeurIPS, ICML, and ICLR), the time from submission deadline to publication is usually 5-6 months. On top of this, open-sourcing of data often involves a delay even when the repository is available at the time of publication~\cite{semmelrock2025reproducibility}. The popularity of pre-print servers such as arXiv decreases this delay~\cite{zhou2025everyone}. Still, there is a gap between the inception of a benchmark to its adoption, which opens the possibility for data contamination and other undesired practices.}, AI models often quickly saturate on new benchmarks, effectively vanishing their discriminatory signal about model performance~\cite{zhou2025lost,srivastava2023beyond}. Proposed solutions include unifying evaluation standards~\cite{bommasani2023holistic}, continuously evaluating the benchmarks themselves~\cite{carro2025conceptual}, or developing fully dynamic benchmarks~\cite{kiela2021dynabench}.

\subsection{Data Contamination and Reliability}
\label{sec:data_contamination}

Data contamination refers to models having seen the benchmark contents during training, effectively allowing them to memorize the data~\cite{deng2024investigating,xu2024benchmark}. To avoid this, strategies utilizing only data from sources published after the AI model's weights were frozen have been proposed~\cite{li2023avoiding}. Overfitting to benchmarks has been demonstrated even in subtle contexts, such as minimal distribution shifts across datasets leading to major performance differences~\cite{zhang2024careful}.

\subsection{Coverage and Discrepancy Between Aims and Measured Signal}

Another known issue with benchmarks is the lack of consistency among different instantiations of benchmarks claiming to test a particular concept, as well as the divergence between the aims of the benchmarks and the actual signal measured. One example of such a domain is reasoning, which suffers from varying definitions and scopes~\cite{fodor2025line,xie2024memorization}, leading to surprisingly poor performance on seemingly trivial tasks~\cite{salido2025none}. Some proposed solutions involve examining coverage through the lens of model activations and interpretability~\cite{bohacek2025uncovering}. Another critique related to the lack of construct validity is the tendency in various AI subfields to prioritize a small number of benchmarks that are treated as milestones towards generalizable AI systems \cite{raji2021ai}.

\subsection{Benchmarking Culture}

\citet{eriksson2025can} examine what they term a ``trust crisis'' in AI evaluation, pointing to construct validity failures and the lack of standardization. Others, including \citet{bliliposition} and \citet{thais2024misrepresented}, examine the narratives and stated research agendas surrounding these benchmarks; some work has found that these patterns differ by region and community~\cite{ott2022mapping}. \citet{weidinger2025toward} have called for a formal ``evaluation science'' for generative AI. Collectively, these unstandardized evaluation practices and their surrounding narratives constitute what \citet{campolo2025state} conceptualizes as a distinct ``benchmarking culture.''

\subsection{AI Benchmarks as Narrative Devices}

Research also shows how AI companies shape the public debate around AI. \citet{nielsen2024how}'s analysis shows that the media coverage of AI "tends to be led by industry sources, and often takes claims about what the technology can and can’t do, and might be able to do in the future, at face value in ways that contributes to the hype cycle." Taking a more nuanced view, \citet{magalhaes2026less}'s qualitative textual analysis of AI coverage in The New York Times (US), De Volkskrant (Netherlands), and Folha de S.Paulo (Brazil) suggests that while journalistic reporting is not necessarily fueling hype, "AI's impact is seen as inevitable but its exact trajectory remains disputed."~\cite{magalhaes2026less}

Others have explored why AI companies dominate public discourse. \citet{khanal2025why} argue that tech monopolies have become "super policy entrepreneurs." They act as "problem brokers" by highlighting certain issues as problem areas, act as "policy entrepreneurs" by providing technical solutions to policy problems, and as "political entrepreneurs" that use their resources to shape political institutions to further their interests."~\cite{khanal2025why} \citet{abdalla2021grey} explored how tech monopolies increasingly influence research through funding to shape the academic expertise governmental bodies  rely on in ways similar to the Big Tobacco industry.

This body of research shows that AI companies shape what counts as state-of-the-art. Benchmarks they choose to highlight are likely to shape public perception despite questions about their scientific validity raised by the work we discussed. In the following, we complement existing literature by analyzing what AI model builders present as state-of-the-art through benchmarks.

\section{Data}

We collect and open-source the \emph{Benchmarking-Cultures-25} dataset, a structured corpus of $231$ unique benchmarks highlighted by $11$ prominent model builders across $139$ distinct generative AI model releases\footnote{For the purposes of this work, we define "generative AI models" as foundation AI models~\cite{bommasani2021opportunities} capable of generating text, code, image, audio, or video in response to input conditioning (most commonly, natural language prompts). We treat all major model variations (e.g., Pro, Flash, Instruct) as distinct releases if separate performance claims were made.} throughout 2025. The dataset is available at \url{https://hf.co/datasets/matybohacek/benchmarking-cultures-25}. Alongside the dataset, we also release an interactive tool to introspect individual benchmarks and explore their relationships to model releases and one another at \url{https://bench-cultures.net} (see Appendix~\ref{app:bench_cultures_screenshots} for screenshots).

To ensure a representative sample of the industry’s state of the art, we selected the top $11$ model builders based on their performance in the LMSYS Chatbot Arena~\cite{chiang2024chatbot} and their inclusion in the "Notable Models" section of the Stanford AI Index 2025~\cite{maslej2025artificial}. This selection captures the dominant organization in the field while maintaining a geographic balance between Western and Chinese organizations. The selected model builders include industry labs (Google, OpenAI, Anthropic, Meta, xAI, Alibaba, Baidu, and DeepSeek) as well as independent and research-oriented organizations (Mistral, Allen Institute for AI, and Z.ai).

\subsection{Data Collection}

For each of the $139$ model releases, we manually extracted every benchmark explicitly mentioned in the primary release announcement; $112$ of these highlighted at least one benchmark. Base models explicitly referenced in announcements were also included. When an announcement covered multiple parameter sizes, we recorded each size as a separate entry. For the purposes of analysis, however, we treated different parameter sizes of the same model as a single release, since model builders vary considerably in how many size variants they publish per model.

Our data collection focused on public-facing industry artifacts (press releases and company blogs) rather than technical documentation (e.g., model cards and API docs) or research papers (e.g., arXiv). To handle variability in how benchmarks are reported, we implemented the following standardization policy:
\begin{itemize}
    \item \textbf{Variant Normalization.} Metric variants (e.g. "HumanEval Pass@1" vs. "HumanEval") were mapped to a single canonical Benchmark ID unless the variation reflected fundamentally different test logic.
    \item \textbf{Snapshot Resolution.} Ambiguous references to dynamic benchmarks (e.g., LiveCodeBench without a date) were resolved using the model's release date and contextual footnotes.
    \item \textbf{Benchmark Author Affiliations.} Using affiliations listed in arXiv papers, the authors of each benchmark were categorized as \texttt{Academic}, \texttt{Industry}, \texttt{Non-profit}, \texttt{Government}, or \texttt{Independent}.
\end{itemize}

To allow for graph analysis of the data, we extended the benchmarks and model releases by a collection of papers, authors, affiliation links, and organizations. In total, we constructed seven data frames with $44$ data fields: Models ($17$), Benchmarks ($6$), Highlights ($4$), Affiliations ($6$), Categories ($3$), and Categorizations ($2$) and Knowledge Subjects ($6$). The complete data structure specification is provided in Appendix~\ref{app:data_structure}.

\subsection{Taxonomy of Tested Competencies}

A core contribution of this study is a unified taxonomy of tested competencies. We inductively extracted what the authors of each benchmark in our dataset claim to measure in their publications and release artifacts (e.g., arXiv paper or Hugging Face repository) and clustered these tested competencies into groups. Through recursive refinement and consensus discussions among the authors, we defined eight meta-categories of tested competencies. A similar process led to the development of additional $22$ categories that break the meta-categories down to more granular capabilities. The complete taxonomy is presented in Appendix~\ref{app:unified_taxonomy}. Once finalized, this taxonomy was used to manually annotate each benchmark recorded in the dataset, unifying the tested competencies. The annotations provide a standardized baseline for comparing how model builders describe benchmarks, who otherwise refer to the same competencies inconsistently. This enables two lines of analysis: first, examining the gaps between how AI model builders frame a benchmark in a release artifact and what the benchmark actually sets out to do; and second, interrogating the construct validity of the benchmarks themselves by comparing their stated aims with what they actually measure.

\subsection{Limitations}

\textbf{Single-year Data Coverage (2025).} We limited our data collection to benchmarks highlighted in model release announcements in 2025 by the selected 11 model builders. This means that our data does not allow us to study broader trends over time, or direct comparisons between publication years.

\textbf{Exclusion of model cards.} We acknowledge that model cards are an important, industry-wide practice to provide more transparency, especially regarding the safety and security of models. However, our study specifically interrogates how model capabilities are advertised to the general public via primary release announcements. We consider our approach as complementary to existing scholarship on the model card landscape.

\textbf{No in-depth analysis of entire benchmark categories.} Analyzing entire benchmark categories qualitatively was beyond the scope of this study. However, we conducted a case study of "General knowledge application" benchmarks limited to the most popular benchmarks of this category (see Section~\ref{sec:case_study}). The analysis provided rich results and illustrates the value of a more comprehensive qualitative analysis.

\textbf{Annotations for own taxonomy done by a single author only.} Multiple independent annotations with inter-rater reliability scoring would have strengthened the classification. To mitigate this limitation, all category assignments were reviewed and discussed among the co-authors, and ambiguous cases were resolved through deliberation.

\section{Data Analysis: Overall Benchmark Origin, Usage, and Presentation Trends}

In this section, we present the overall statistics and trends in the \emph{Benchmarking-Cultures-25} dataset by examining the use of benchmarks in $139$ models released in 2025 from $11$ AI model builders. Out of these, $35$ models are closed-source, $94$ are open-weight, and $10$ are fully open-source. Four model builders in our dataset are Chinese (Alibaba, Baidu, DeepSeek and Z.ai); the remaining seven are US- or Europe-based (Allen Institute for AI, Anthropic, Google, Meta, Mistral, OpenAI, and xAI).


\subsection{Benchmark Origin (RQ1)}

\textbf{Increasingly, benchmarks highlighted in model release artifacts are published by industry rather than academia.} $43.9\%$ of the benchmark authors are affiliated with industry, $39.0\%$ with academia. These numbers are more pronounced for Western model builders, where the number of benchmark authors affiliated with industry is $52.3\%$. Authors of benchmarks published in 2025 have an even higher industry affiliation rate. This trend is, too, more pronounced for Western model builders, where this was $64.5\%$ (see Table~\ref{tab:bench_author_affil}).

\begin{table*}[t]
\begin{minipage}[t]{0.58\textwidth}
\raggedright
\caption{\textbf{Affiliation of Benchmark Authors.} Authors' affiliations are categorized into organization categories. A breakdown is provided for benchmarks published in 2025 and all benchmarks present in the dataset.}
\label{tab:bench_author_affil}
\begin{tabular}{l cc cc cc}
\toprule
\textbf{Organization} & \multicolumn{2}{c}{\textbf{Overall (\%)}} & \multicolumn{2}{c}{\textbf{West (\%)}} & \multicolumn{2}{c}{\textbf{China (\%)}} \\
\cmidrule(r){2-3} \cmidrule(lr){4-5} \cmidrule(l){6-7}
\textbf{Category} & All & 2025 & All & 2025 & All & 2025 \\
\midrule
Industry & 43.9 & 49.3 & 52.3 & 64.5 & 42.6 & 46.2 \\[1pt]
Academia & 39.0 & 33.0 & 38.2 & 27.9 & 42.8 & 45.4 \\[1pt]
Non-profit & 10.1 & 6.8 & 9.1 & 7.4 & 14.7 & 8.3 \\[1pt]
Other/Unspec. & 5.7 & 9.6 & -- & -- & -- & -- \\[1pt]
Independent & 0.7 & 1.3 & -- & -- & -- & -- \\[1pt]
Government & 0.4 & 0.1 & 0.4 & 0.2 & -- & -- \\[1pt]
\bottomrule
\end{tabular}
\end{minipage}
\hfill
\begin{minipage}[t]{0.38\textwidth}
\raggedright
\caption{\textbf{Benchmark Authors with Multiple Affiliations.} Distribution of affiliation combinations among authors with multiple affiliation.}
\label{tab:author_multiple_affiliations}
\begin{tabular}{lc}
\toprule
\textbf{Affiliation} & \textbf{Authors} \\
\textbf{Combination} & \textbf{(\%)} \\
\midrule
Academia \& Non-profit & 37.1 \\
Academia \& Industry & 21.8 \\
Academia \& Academia & 19.8 \\
Industry \& Industry & 9.6 \\
Academia \& Government & 5.1 \\
Industry \& Non-profit & 4.6 \\
Acad. \& Ind. \& Non-profit & 1.5 \\
\bottomrule
\end{tabular}
\end{minipage}
\end{table*}

\textbf{$\textbf{8.1\%}$ ($\textbf{198}$) of benchmark authors have more than one affiliation.} Of these, $37.1\%$ have an affiliation with a non-profit as well as with academia. This derives from the large contribution by authors who are affiliated with the Allen Institute for AI, which usually have an additional academic affiliation. Almost a third of those $197$ authors ($32.9\%$) have a shared affiliation between industry and some other type of organization (see Table~\ref{tab:author_multiple_affiliations}).

\subsection{Presentation of Tested Competencies (RQ2)}

Reported in Table~\ref{tab:top15_bench_popularity_cat} are the competencies tested in the top 15 most popular benchmarks. We see that $41.7\%$ of them evaluate "Math", followed by "Reasoning and knowledge" (i.e., reasoning in fields other than math or coding) with $25.0\%$. Notably, all 15 top benchmarks that evaluate "Reasoning and knowledge" also include math as a subject, hence the overlap of benchmarks in Table~\ref{tab:top15_bench_popularity_cat} (in our own taxonomy, each benchmark could be assigned to multiple categories to reflect overlaps such as this one).

\begin{table}[b]
\centering
\caption{\textbf{Distribution of Evaluated Competencies in the Top 15 Most Popular Benchmarks.} All benchmarks in the "Reasoning and knowledge" category are also used to evaluate "Math" competency. Hence, they are listed twice. Listed competencies are based on our own taxonomy.}
\label{tab:top15_bench_popularity_cat}
\begin{tabular}{l c p{8cm}}
\toprule
\textbf{Tested Competency} & \textbf{Highlights (\%)} & \textbf{Benchmarks} \\
\midrule
Math & 41.7 & AIME 2024, AIME 2025, GPQA Diamond, HMMT 2025, Humanity's Last Exam, MATH, MMLU, MMLU-Pro, MMMLU, MMMU \\
\addlinespace
Reasoning and knowledge & 25.0 & GPQA Diamond, Humanity's Last Exam, MMLU, MMLU-Pro, MMMLU, MMMU \\
\addlinespace
Coding & 12.5 & Aider Polyglot, LiveCodeBench, SWE-bench Verified \\
\addlinespace
Audio Visual Understanding & 8.3 & Humanity's Last Exam, MMMU \\
\addlinespace
Factuality & 4.2 & SimpleQA \\
\addlinespace
Instruction & 4.2 & IF-Eval \\
\addlinespace
Multilingual Performance & 4.2 & MMMLU \\
\bottomrule
\end{tabular}
\end{table}

\begin{figure}
    \centering
    \includegraphics[width=0.9\linewidth]{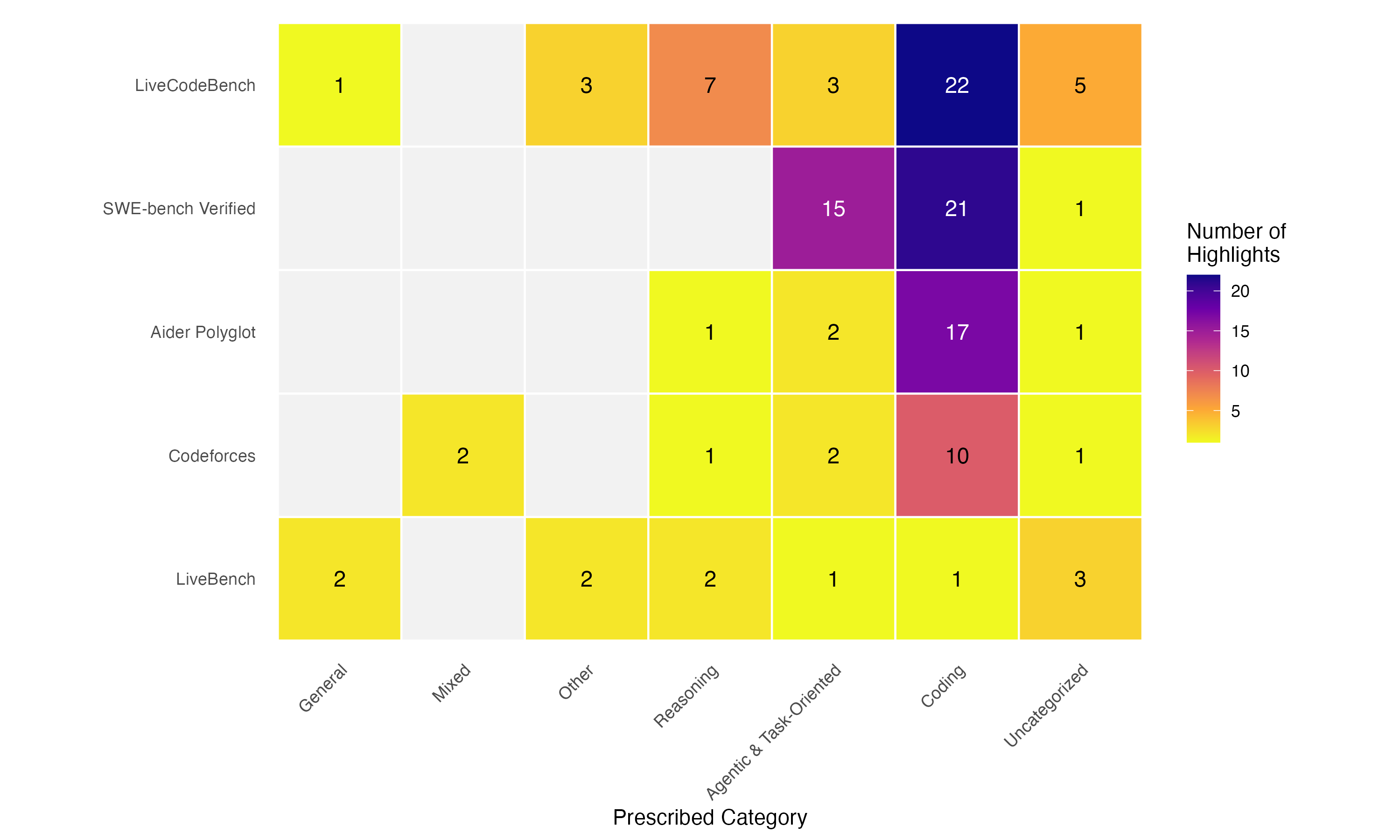}
    \caption{\textbf{Prescribed Competencies by Model Builders Within The Top 5 "Coding" Benchmarks.} This graph shows the count of competency categories that model publishers prescribe to benchmarks across model releases.}
    \label{fig:coding_bench_cat}
    \vspace{-0.6em}
\end{figure}

\textbf{Model builders inconsistently label the same benchmarks to represent different competencies across releases, even between model releases by the same organization.} Shown in Figure~\ref{fig:coding_bench_cat} are the types of labels that model builders used to describe tested competencies by benchmarks across model releases, indicating that model builders are inconsistent in how they frame benchmarks. LiveCodeBench, the third most popular benchmark in our dataset overall, is a good example to illustrate this. The authors of LiveCodeBench describe it as "a holistic and contamination-free benchmark for evaluating code capabilities." \cite{jain2024livecodebench} We therefore categorized it as \textit{Specialized knowledge application - Coding} in our taxonomy. However, only $53.7\%$ of model release artifacts presented LiveCodeBench as a coding-related benchmark. Some model builders refer to it as "Reasoning" (DeepSeek, Mistral, and Z.ai) or agent-related functions (Z.ai and DeepSeek). What LiveCodeBench is claimed to evaluate is even inconsistent between model releases by the same model builder. For example, xAI pivoted from "Coding" to "Cost-efficient Intelligence," and Alibaba presents it either as evaluating instructions, "post-training" or simply "text." We found similar inconsistencies across all benchmarks.

\subsection{Benchmark Popularity (RQ3)}
\label{subsec:bench_popularity}

We ranked benchmarks by popularity using the geometric mean $\sqrt{N_{\text{builders}} \cdot N_{\text{highlights}}}$ of the number of model builders and the number of model releases highlighting each benchmark. A simple highlight count would be skewed by the uneven release volumes across the $11$ model builders, risking overrepresentation of benchmarks favored by high-volume model builders. We use the geometric mean to normalize benchmark prevalence, yielding rankings that better reflect broad, cross-industry adoption rather than the idiosyncrasies or release frequency of individual model builders. The results are shown in Table~\ref{tab:top15_bench_popularity}.

\newcolumntype{C}[1]{>{\centering\arraybackslash}p{#1}}

\begin{table}[h]
\centering
\caption{\textbf{Top 15 Most Popular Benchmarks.} Benchmarks are ranked by popularity score (see Section~\ref{subsec:bench_popularity}).}
\label{tab:top15_bench_popularity}
\begin{tabular}{l c C{1.45cm} C{1.45cm} c c c}
\toprule
 & & \multicolumn{2}{c}{\textbf{Model Builders (Rank)}} & \multicolumn{3}{c}{\textbf{Models (Rank)}} \\
\cmidrule(lr){3-4} \cmidrule(lr){5-7}
\textbf{Benchmark} & Overall Rank & \centering West & \centering China & Closed & Open-Weight & Open-Source \\
\midrule
AIME 2025 & 1 & 3 & 3 & 3 & 3 & 5 \\
GPQA Diamond & 2 & 1 & 8 & 1 & 4 & - \\
LiveCodeBench & 3 & 6 & 1 & 14 & 1 & 12 \\
MMLU-Pro & 4 & 7 & 2 & 16 & 2 & 12 \\
AIME 2024 & 5 & 4 & 4 & 5 & 5 & 12 \\
MMMU & 6 & 2 & 30 & 2 & 11 & - \\
SWE-bench Verified & 7 & 5 & 5 & 4 & 7 & - \\
SimpleQA & 8 & 8 & 6 & 7 & 8 & 23 \\
Humanity's Last Exam & 9 & 12 & 8 & 6 & 10 & - \\
MMLU & 10 & 14 & 8 & 35 & 6 & 1 \\
MATH & 11 & 11 & 12 & 24 & 11 & 2 \\
MMMLU & 11 & 8 & 45 & 8 & 19 & - \\
Aider Polyglot & 13 & 10 & 20 & 8 & 22 & - \\
IF-Eval & 13 & 13 & 12 & 24 & 13 & 2 \\
HMMT 2025 & 15 & 20 & 11 & 18 & 13 & - \\
\bottomrule
\end{tabular}
\vspace{-1em}
\end{table}

\textbf{AIME 2025 is the most popular benchmark overall, closely followed by GPQA Diamond and LiveCodeBench.} Notably, GPQA Diamond is more popular with model builders from the West (ranking 1st) than from China (ranking 8th). LiveCodeBench is more popular among open-weight and open-source models (ranking 1st and 12th, respectively) than proprietary models, where it ranks as the 14th most popular. 


\subsection{Adoption and Cross-Model Comparability (RQ4)}

To get a sense of how quickly model builders start highlighting benchmarks after their first release, we calculated the adoption rate as the number of model release announcements who have highlighted a benchmark since its release.

\begin{figure}[t]
    \centering
    \includegraphics[width=0.9\linewidth]{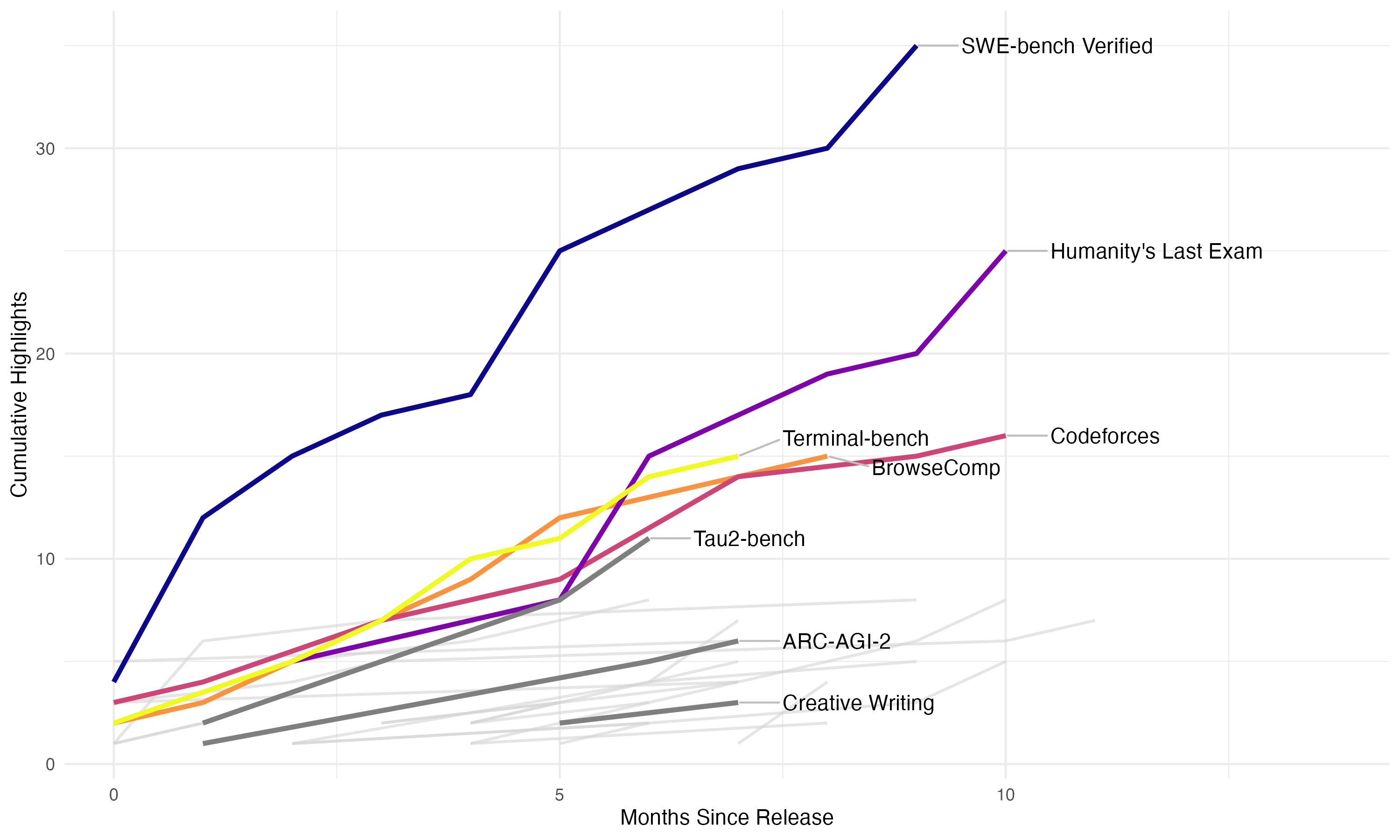}
    \caption{\textbf{Adoption of Benchmarks Released in 2025.} The top five most adopted models are highlighted for clarity.}
    \label{fig:bench_adoption_rate}
\end{figure}

\textbf{$\textbf{71.9\%}$ of benchmarks used in 2025 were published in the last three years.} The cumulative adoption rate of the benchmarks published in 2025 is shown in Figure~\ref{fig:bench_adoption_rate}. The majority ($31.6\%$) were published in 2024, followed by $28.1\%$ in 2025. SWE-bench Verified was by far the most adopted benchmark of all benchmarks published in 2025, followed by Humanity's Last Exam (HLE). This makes SWE-bench Verified the seventh and HLE the ninth most popular benchmark. For closed models, SWE-bench Verified and HLE is even more popular and take the fourth and the sixth rank, respectively.

\begin{table}[t]
\vspace{-0.2em}
\centering
\caption{\textbf{Publication Years of Benchmark within Selected Tested Competencies.} Looking at the benchmarks released in 2023, 2024, and 2025 we map the number of benchmarks released per year within a tested competency. See Table~\ref{tab:all_competencies} in Appendix~\ref{app:full_tables} for full data.}
\label{tab:selected_competencies}
\small
\begin{tabular}{llcccccc}
\toprule
& & \multicolumn{2}{c}{\textbf{2023}} & \multicolumn{2}{c}{\textbf{2024}} & \multicolumn{2}{c}{\textbf{2025}} \\
\cmidrule(lr){3-4} \cmidrule(lr){5-6} \cmidrule(lr){7-8}
\textbf{Meta-Category} & \textbf{Category} & $N$ & \% & $N$ & \% & $N$ & \% \\
\midrule
Agentic task execution & Strategic planning & - & - & - & - & 12 & 100.0 \\
Agentic task execution & Tool orchestration & - & - & 5 & 38.5 & 7 & 53.8 \\
General knowledge application & Reasoning and knowledge & 10 & 22.2 & 12 & 26.7 & 9 & 20.0 \\
Specialized knowledge application & Coding & 3 & 12.0 & 9 & 36.0 & 7 & 28.0 \\
Specialized knowledge application & Math & 11 & 21.6 & 14 & 27.5 & 8 & 15.7 \\
\bottomrule
\end{tabular}
\end{table}

\begin{figure}
    \centering
    \includegraphics[width=0.9\linewidth]{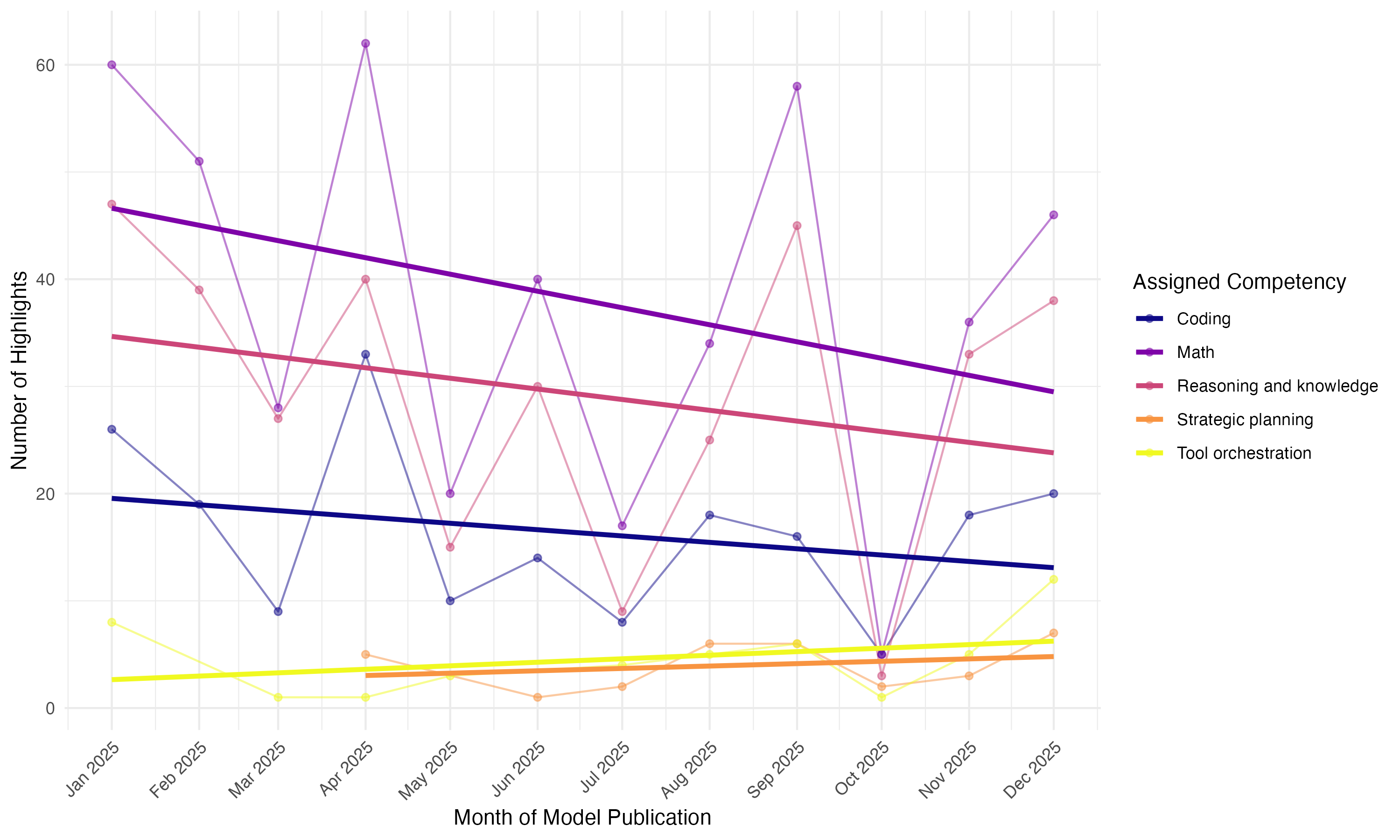}
    \caption{\textbf{Highlight Frequency of Selected Competencies by Model Builders.} This graph shows the trend of these selected competencies being highlighted in model releases throughout 2025. See Figure~\ref{fig:highlights} in Appendix~\ref{app:full_tables} for the full graph.}
    \label{fig:selected_highlights}
\end{figure}

\textbf{Adoption for new benchmarks show a trend towards more "Agentic task execution" benchmarks.} When we look at the release dates of benchmarks highlighted by model builders in 2025, we can identify a few trends (see Table~\ref{tab:selected_competencies}). For some competencies, model builders tend to rely more on older benchmarks. Models highlighted $6.7\%$ fewer benchmarks for "Reasoning and knowledge" released in 2025 over benchmarks released in 2024. A similar decrease can be observed for benchmarks testing for "Coding" ($8.0\%$) or "Math" ($11.8\%$). We also see competencies that are novel in 2025 and were quickly adopted by model builders, most importantly agentic competencies such as "Strategic Planning" and "Tool orchestration," or very recently also preference alignment for specific domains like "Health." This dynamic is also reflected in the model releases and the competencies they choose to highlight, as seen in Figure~\ref{fig:selected_highlights}. Again, "Math", "Coding," and "Reasoning and knowledge" saw a decline in inclusion in release artifacts, while competencies around agentic capabilities saw a steady increase.

\textbf{Benchmark selection is highly fragmented, limiting cross-model comparability.} Table~\ref{tab:bench_usage_affil_distrib} shows how frequently benchmarks are highlighted across models. $63.2\%$ of the benchmarks ($146$) are used only by a single model builder. There are differences between the West and China, where respectively $70.3\%$ and $64.7\%$ of all benchmarks were used by a single model builder. $89$ benchmarks ($38.5\%$) are used by a single model. $51.3\%$ of closed models ($39$ in total) reuse benchmarks three or fewer times. This number is even higher for open-weight models, where more than $66.5\%$ of all models reuse the same benchmark three times or less.

\begin{table}[h]
\centering
\caption{\textbf{Distribution of Benchmark Adoption.} Percentage of model builders and models that include a given benchmark exactly $N$ times across their release artifacts.}
\label{tab:bench_usage_affil_distrib}
\begin{tabular}{c ccc cccc}
\toprule
\textbf{$N$} & \multicolumn{3}{c}{\textbf{Model Builders (\%)}} & \multicolumn{4}{c}{\textbf{Models (\%)}} \\
\cmidrule(r){2-4} \cmidrule(l){5-8}
\textbf{(\# Highlights)} & Total & West & China & Total & Closed & Open-Weight & Open-Source \\
\midrule
1 & 63.2 & 70.3 & 64.7 & 38.5 & 23.7 & 39.3 & 55.1 \\
2 & 13.0 & 11.7 & 22.2 & 13.4 & 17.1 & 19.1 & 22.4 \\
3 & 10.0 & 7.8 & 10.8 & 12.6 & 10.5 & 8.1 & 10.2 \\
4 & 4.8 & 3.9 & 2.4 & 7.4 & 9.2 & 9.8 & 4.1 \\
5 & 1.7 & 2.3 & - & 6.9 & 5.3 & 5.2 & 6.1 \\
\bottomrule
\end{tabular}
\end{table}

Looking at specific benchmarks, AIME 2025 was highlighted most frequently (in $46.8\%$ of the analyzed model release artifacts). From there, the frequencies of individual benchmark decrease steeply: MMMLU, the tenth most highlighted benchmark, only appears in $24.5\%$ of the analyzed model releases artifacts, and HMMT 2025, the 15th most highlighted benchmark, only in 16.0\%.

\section{Case Study: General knowledge application}
\label{sec:case_study}


Some types of comprehension and reasoning, such as math and coding, can utilize existing real-world resources (like annual math competitions), including prescribed languages and testing procedures, and their evaluation is, hence, largely standardized. Benchmarks measuring "General knowledge application", however, are more ambiguous because they evaluate knowledge retrieval, comprehension, or reasoning across a broad spectrum of disciplines, ranging from STEM to the humanities, law and more. Despite the ambiguity, "General knowledge application" represents the second most popular benchmark category in our dataset: $74.5\%$ of all model release announcements highlighted at least one of the top five "General knowledge application" benchmarks. Given this combination of popularity and difficulty of evaluation, we analyzed these top five benchmarks in depth to better understand what they, as the most frequently highlighted benchmarks in our dataset, measure and how consistent they are in their stated goals.\footnote{We excluded MMMLU from our analysis despite being in the top five, since it is a translation of MMLU's test set, which is already included.} The analysis that follows focuses on these five benchmarks specifically, not the category as a whole. 

\begin{table}[t]
    \centering
    \small 
    \renewcommand{\arraystretch}{1.3} 
    \caption{\textbf{Stated Goals and Subject Coverage of the Top Five "General Knowledge Application" Benchmarks.} Despite claiming to measure general knowledge or reasoning broadly, all five benchmarks focus heavily on STEM subjects.}
    \label{tab:top5_reasoning_bench}
    \begin{tabular}{p{2cm} p{4.5cm} p{7.5cm}} 
        \hline
        \textbf{Benchmark} & \textbf{Stated Goal} & \textbf{Subjects covered} \\
        \hline
        GPQA Diamond & Help to develop "scalable oversight" methods, i.e. enable experiments in scenarios where model output is hard to verify for experts. & Chemistry (47\%), Physics (43\%), Biology (10\%)\\
        \hline
        MMMU & Evaluate multimodal models on "massive" multi-discipline tasks demanding college-level subject knowledge and "deliberate reasoning." & Engineering (26\%), Science (23\%), Medicine (17\%), Business (14\%), Art \& Design (11\%), Humanities \& Social Science (9\%) \\
        \hline
        MMLU-Pro & Extend the "mostly knowledge-driven" MMLU benchmark by integrating more "reasoning-focused" questions. & Math (11\%), Physics (11\%), Chemistry (9\%), Law (9\%), Engineering (8\%), Other (8\%), Economics (7\%), Health (7\%), Psychology (7\%), Business (7\%), Biology (6\%), Philosophy (4\%), Computer Science (3\%), History (3\%) \\
        \hline
        MMLU & Comprehensively measure the knowledge models have across multiple domains. & STEM (33\%), Other (23\%), Humanities (23\%), Social Sciences (21\%) \\
        \hline
        Humanity's Last Exam & Performance on closed-ended questions regarding "cutting-edge" scientific knowledge. & Math (41\%), Biology/Medicine (11\%), Computer Science/Artificial Intelligence (10\%), Physics (9\%), Humanities/Social Science (9\%), Other (9\%), Chemistry (7\%), Engineering (4\%) \\
        \hline
    \end{tabular}
\end{table}

As illustrated in Table~\ref{tab:top5_reasoning_bench}, the stated goals of these examined benchmarks are broad, as they claim to measure general knowledge or reasoning, despite focusing only on select subjects and not covering various other subjects systematically or equally. A deeper look into these benchmarks reveals several key implications for the benchmarking cultures of AI model builders.

\textbf{Breakdown of tested subjects.} We break down the subjects benchmark authors claim to cover in "Knowledge and reasoning" benchmarks in Table~\ref{tab:reasoning_knowledge_tables}. "Science" is by far the most popular field with almost a third of all questions relating to it, followed by "Humanities \& Social Sciences" with almost half the amount of questions. Trailing behind is "Art \& Design". A closer look at the science category reveals a strong imbalance within the different sub fields, with more than a third of all questions related to mathematics. This is additional to the dedicated evaluations for math.

\textbf{All top five "General knowledge application" benchmarks distinguish between knowledge and reasoning, but do not define what the distinction is.} MMLU from 2020 is the oldest benchmark in Table~\ref{tab:top5_reasoning_bench} and the only one with an emphasis on knowledge. The authors argued that previous benchmarks in Natural Language Processing (NLP) evaluated linguistic skills, but MMLU should evaluate information contained in model's pretraining data, which the authors refer to as "knowledge": "To bridge the gap between the wide-ranging knowledge that models see during pretraining and the existing measures of success, we introduce a new benchmark for assessing models across a diverse set of subjects that humans learn."~\cite{hendrycks2020measuring} Essentially, the benchmark is meant to evaluate not only what information was contained in the pretraining data, but also how well models are able to recall it correctly when prompted. "Reasoning" is only mentioned in relation to the subjects covered, which would require various forms of reasoning. Implicitly, reasoning thus appears to be understood as "applying" knowledge from pretraining to solve tasks: "We introduced a new test that measures how well text models can learn and apply knowledge encountered during pretraining."~\cite{hendrycks2020measuring}.

\begin{figure}
    \centering
    \includegraphics[width=1\linewidth]{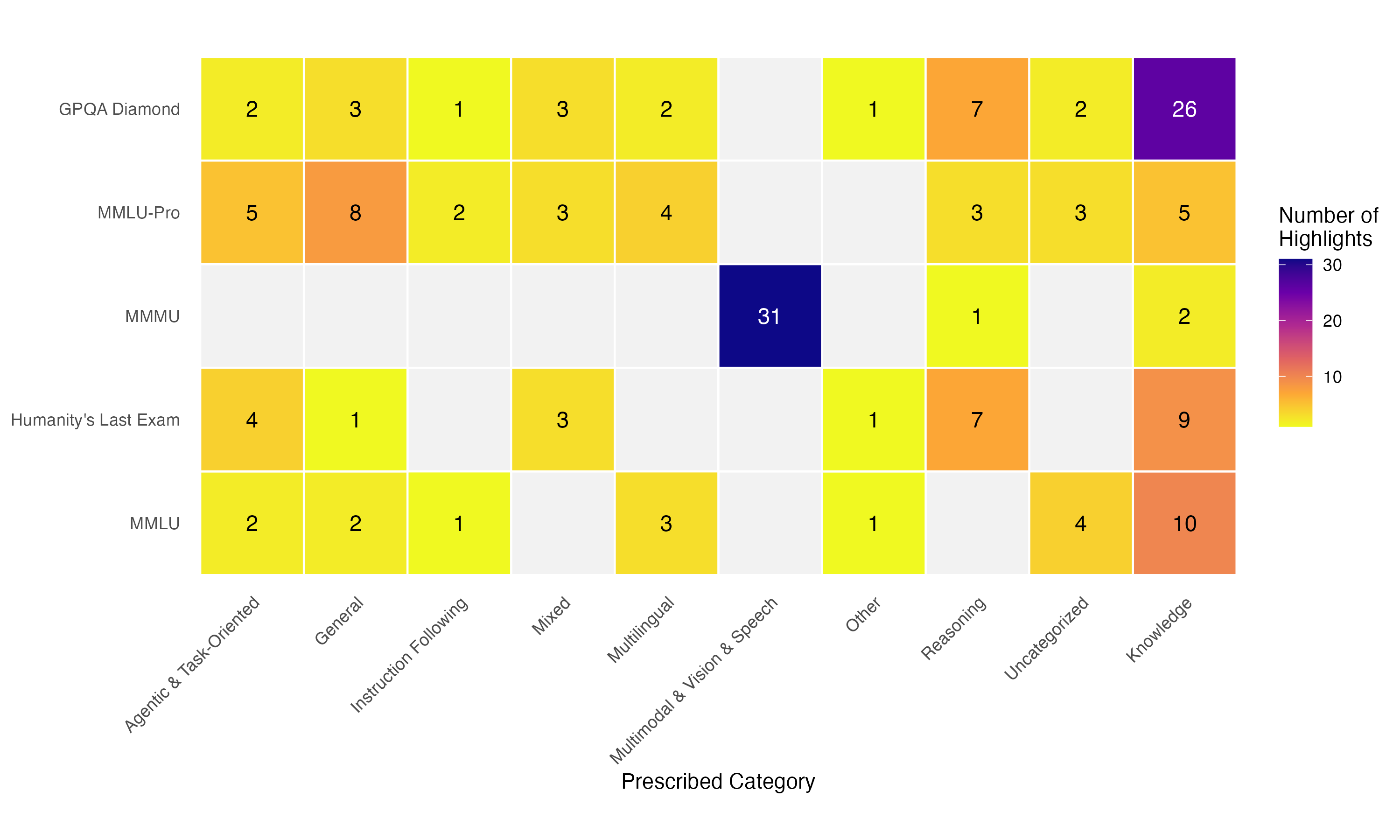}
    \caption{\textbf{Prescribed Competencies by Model Builders Within The Top Five "Reasoning and knowledge" Benchmarks.} This heatmap shows the count of competency categories that model builders prescribe to benchmarks across model releases. MMMLU is excluded as it is a translation of MMLU's test set.}
    \label{fig:top5_reasoning_cat}
\end{figure}

\textbf{All top five "General knowledge application" benchmarks but MMLU emphasize reasoning over knowledge, which they implicitly define as making logical inferences.} MMLU-Pro is supposed to "extend the mostly knowledge-driven MMLU benchmark by integrating more challenging, reasoning-focused questions."~\cite{wang2024mmlupro} In practice, MMLU-Pro added six incorrect but plausible options to multiple-choice questions and increased the number of college-level exam problems that would require "deliberate reasoning,"~\cite{wang2024mmlupro}, a term that the authors do not further define in their paper. The clearest indication of what the authors understand as "reasoning" is their error analysis of GPT-4o: "The model frequently encounters difficulties with logical reasoning, even when it recalls the correct information and knowledge"~\cite{wang2024mmlupro}. By implication, reasoning is understood to be the making of logical inferences. The authors of the MMMU benchmark similarly describe "reasoning errors" as errors "where the model correctly interprets text and images and recalls relevant knowledge... [yet] fails to apply logical and mathematical reasoning skills effectively to derive accurate inferences"~\cite{yue2024mmmu}.

\textbf{To ensure reasoning is required, authors of reasoning-focused benchmarks claim to develop tasks that are "non-searchable".} GPQA Diamond and HLE are less explicit about their understanding of reasoning but use similar metaphors as the authors of MMLU-Pro and MMMU. The questions in both GPQA Diamond and HLE should be "non-searchable." GPQA's questions were designed to have a ground truth known to experts, but not to "non-experts using easily-found internet resources, since we require that questions be hard and Google-proof in order to be suitable for scalable oversight experiment"~\cite{rein2023gpqa}. For HLE, questions "should be precise, unambiguous, solvable, and non-searchable, ensuring models cannot rely on memorization or simple retrieval methods"~\cite{phan2025humanitys}. Moreover, the HLE authors put an emphasis on mathematics problems "aimed at testing deep reasoning skills broadly applicable across multiple academic areas" \cite{phan2025humanitys}.

\textbf{However, what is missing in the reasoning-focused benchmarks in Table~\ref{tab:top5_reasoning_bench} is a reflection about the extent to which models really rely on logical inference rather than anything akin to what authors consider "knowledge" to solve tasks.} As mentioned in Section~\ref{sec:data_contamination}, data contamination is well-known issue in benchmarking, which skews models towards relying on knowledge rather than reasoning. Likewise, arguing that reasoning tasks are difficult because they are non-searchable arguably conflates information scarcity with the complexity or difficulty of the task. There is also the implicit assumption that reasoning happens on a scale: HLE questions should not just test reasoning, but "deep reasoning"~\cite{phan2025humanitys}, the authors of MMLU-Pro make a distinction between "reasoning-focused" subjects (like math or physics) and "knowledge-heavy" ones (like history or law)~\cite{wang2024mmlupro}. Implicitly, "more" or "deeper" reasoning is tied to questions that require more specialist domain expertise, while lower levels or reasoning are associated with common sense questions. However, these assumptions are not made explicit and are not examined. In addition, benchmark authors talk about measuring knowledge and reasoning in broad and general terms. For example, the authors of MMMU argue that they measure progress towards AI systems that equal "at least 90th percentile of skilled adults in a broad range of tasks"~\cite{yue2024mmmu}. The authors of GPQA claim to evaluate on tasks that border "the frontier of human knowledge" \cite{rein2023gpqa}.

\textbf{This lack of construct validity reflection appears to be partly driven by some benchmark authors' goal of measuring progress towards AGI.} The authors of MMMU and MMLU-Pro explicitly aim to help measure progress towards AGI following a framework defined by \citet{morris2024levels}. The framework consists of five "Levels of AGI" based on the performance and generality of AI systems. Following \citet{morris2024levels}, knowledge and reasoning are essential to progress to higher AGI levels: "The ability to learn new skills...is essential to generality, since it is infeasible for a system to be optimized for all possible use cases a priori; this necessitates related sub-skills such as the ability to select appropriate strategies for learning" \cite{morris2024levels}. The authors of MMMU and MMLU-Pro both specifically want to measure progress towards what \citet{morris2024levels} call "Expert AGI:" an AI system that reaches "at least 90th percentile of skilled adults" on a "wide range of non-physical tasks." It is only the third level in their framework, but reaching it, they argue, would likely cause economic disruption as it would enable industries to "reach the substitution threshold for machine intelligence in lieu of human labor" \cite{morris2024levels}. Therefore, the authors of MMMU argue "it is of both intellectual and societal importance to closely monitor the progress towards Expert AGI." \cite{yue2024mmmu}

\textbf{However, the main inspiration of the MMMU and MMLU-Pro authors~\cite{morris2024levels} remains vague about how progression towards various levels of AGI should be measured.}  What constitutes the 90th percentile of "skilled adults"? And on how many tasks should an AI system reach their performance to cover "most" tasks these skilled adults can perform? \cite{morris2024levels} broadly suggest that an "AGI benchmark" should evaluate a model's "ability to learn new skills...the ability to know when to ask for help, and... social metacognitive abilities such as those relating to theory of mind." Subsequently, the authors of MMMU and MMLU-Pro emphasize reasoning over knowledge and highlight the broad range of tasks and subjects covered by their benchmarks. This might be sufficient to claim to help measure progress towards "Expert AGI" as defined by \cite{morris2024levels}, but the questions about construct validity raised above remain.

\textbf{We also found that GPQA Diamond and HLE are clearly informed by AGI narratives without explicitly citing AGI frameworks.} The GPQA Diamond authors caution that if "narrowly superhuman AI systems could help to advance the frontier of human knowledge," they are likely to produce answers that are difficult to verify even for subject-matter experts \cite{rein2023gpqa}. Their goal is to support experiments with "scalable oversight," a concept introduced by~\citet{amodei2016concrete}. The authors of HLE claim to evaluate the "frontier of human knowledge, designed to be the final closed-ended academic benchmark of its kind with broad subject coverage" \cite{phan2025humanitys}. In this vein, it is only fitting that its authors originally planned to name their benchmark "Humanity's Last Stand."\cite{roose2025WhenAIPassesThisTest} Branding a benchmark as "final" or as evaluating "frontier knowledge" implies a teleological inevitability about AGI. The authors also stress that good performance on HLE "would not alone suggest autonomous research capabilities or 'artificial general intelligence'" \cite{phan2025humanitys}. This mirrors \citet{morris2024levels}'s language about the importance of AGI systems to learn new skills to achieve generality.

\begin{table}[t]
\noindent
\begin{minipage}[t]{0.48\linewidth}

\centering
\captionof{table}{Distribution of Subjects covered in Top 5 (excluding MMMLU) "Reasoning and knowledge" Benchmarks by Field.}
\label{tab:reasoning_knowledge_tables}
\begin{tabular}{lcc}
\toprule
& \multicolumn{2}{c}{\textbf{Share}} \\
\cmidrule(lr){2-3}
\textbf{Field} & $N$ & \% \\
\midrule
Science & 12,850 & 30.4 \\
Humanities \& Social Sciences & 7,920 & 18.8 \\
Business & 5,095 & 12.1 \\
Tech \& Engineering & 4,877 & 11.6 \\
Health \& Medicine & 4,872 & 11.5 \\
Law & 3,078 & 7.3 \\
Other/Unspecified & 2,187 & 5.2 \\
Art \& Design & 1,329 & 3.1 \\
\bottomrule
\end{tabular}
\end{minipage}
\hfill
\begin{minipage}[t]{0.48\linewidth}
\centering
\captionof{table}{Breakdown of Disciplines covered in the Science Field in "Reasoning and knowledge" Benchmarks.}
\label{tab:field_distrib_science}
\begin{tabular}{l cc}
\toprule
& \multicolumn{2}{c}{\textbf{Share}} \\
\cmidrule(lr){2-3}
\textbf{Discipline} & $N$ & \% \\
\midrule
Math & 4,395 & 34.2 \\
Physics & 3,016 & 23.5 \\
Chemistry & 2,353 & 18.3 \\
Biology & 2,207 & 17.2 \\
Geography & 879 & 6.8 \\
\bottomrule
\end{tabular}

\end{minipage}
\end{table}

\section{Discussion}

\textbf{The way model builders highlight benchmark results only offers very limited cross-modal comparison.} Model builders are very inconsistent about the benchmarks they highlight and how they frame them. Our analysis of the top five benchmarks evaluating "General knowledge application" illustrates that among the few benchmarks that are used more widely, several put an emphasis on measuring progress towards vaguely defined concepts of AGI over construct validity, which further undermines model comparison. 

\textbf{Criticism about the quality of a benchmark does not seem to have much impact on its popularity among model builders.} Despite their popularity, several benchmarks in Table~\ref{tab:top5_reasoning_bench} have been shown to contain incorrect information. In July 2025, FutureHouse published a review of HLE pointing out "that 29 ± 3.7\% (95\% CI) of the text-only chemistry and biology questions had answers with directly conflicting evidence in peer reviewed literature."~\cite{white2025ChemistryBiologyLikelyWrong} However, more than 60\% of all mentions of HLE in model release artifacts appeared after FutureHouse's publication. Uncertainty about the veracity of some of the contents of HLE did not stop its adoption by AI model builders. As mentioned in our discussion of related work above, MMLU has also been criticized for containing a substantial amount of errors, including wrong ground truths \cite{gema2025we}.

\textbf{When presenting general purpose models, model builders in our dataset frequently imply their model's potential to replace human labor with their selection of benchmarks.}  When model builders prominently highlight increased performance on benchmarks that explicitly or implicitly aim to track progress towards AGI they imply that their model is getting closer to AGI and thus has a bigger capacity to replace human labor. GPQA Diamond is worth pointing out here as the most frequently highlighted benchmark in our data. Its stated goal is not to evaluate specific model capabilities but to help develop methods to verify the correctness of a model's response in scenarios where even subject-matter experts struggle to verify it. A high score of GPQA Diamond thus suggests that a model is potentially "dangerous" because its capabilities have outpaced human oversight mechanisms, feeding into the narrative of creating "superhuman AI systems."

\textbf{We also found a decline in independent benchmarks being highlighted by model builders.} Increasingly, benchmark authors are affiliated with industry rather than academia. Model builders also increasingly highlight benchmarks they created themselves. Especially OpenAI highlighted 10 benchmarks it created itself. A total of 36 benchmarks were fully or partly created by the model builders that evaluated one of their own models against it. This trend is increasing, with 52.8\% of these benchmarks being published in 2025.

\textbf{Model builders focus on performance while leaving safety concerns unaddressed.} In public debate, there are many concerns about the biases, potential harms, and safety issues of generative AI models. Yet, not a single benchmark in our dataset addresses these issues. For example, there was no benchmark evaluating robustness against prompt injection, or that evaluated how race and gender tend to be framed by a model. Those issues are typically reserved to model cards, but those are less public-facing than public model release announcements.

\textbf{Benchmarks serve as narrative devices.} We observed several trends that show a change in the way benchmarks are created and used. Increasingly, (1) benchmarks are produced by authors in the industry, (2) benchmarks are created by model builders with the purpose of evaluating their own models, and (3) we see a shift in tested competencies that align with broader narratives around generative AI models and AGI. Benchmarks increasingly serve a dual purpose: they are marketing tools as much as they serve a scientific process. The boundaries between the two are murky and, looking at benchmarks published in 2025, increasingly disappearing. Benchmarks highlighted by model builders often say less about the real performance of their AI models and more about their aspirations.

\newpage

\section*{Author Contributions}
SB, CB, and MB jointly developed the methodology, conducted the data analysis, and wrote the paper. SB and CB led the data collection effort. CB additionally designed and built the accompanying interactive tool.

\section*{Generative AI Usage Statement}
Generative AI tools were used for literature search, proofreading, LaTeX table and figure formatting, and grammatical corrections. They were not used during data collection, normalization, or annotation, all of which were conducted manually by the authors.

\section*{Acknowledgments}
CB thanks the Mozilla Foundation for its support during the fellowship over which this work was conducted.

\section*{Competing Interests}
MB was previously employed by Google DeepMind, which is among the model builders whose benchmarking practices are analyzed in this paper. The analysis, findings, and conclusions are the authors' own and do not reflect the views of Google DeepMind. SB and CB declare no competing interests.

\section*{Ethical Considerations Statement}
This research did not involve human subjects, collection of private data, or interventions. The released dataset consists of openly available metadata with links and attribution, and does not redistribute proprietary content.

\newpage

\bibliographystyle{ACM-Reference-Format}
\bibliography{refs}

\newpage
\onecolumn
\appendix

\section{\emph{Benchmarking-Cultures-25} Data Structure}
\label{app:data_structure}

This section describes the complete core data structure of our \emph{Benchmarking-Cultures-25} dataset, which consists of seven data frames with a total of $44$ data fields: Models ($17$), Benchmarks ($6$), Highlights ($4$), Affiliations ($6$), Categories ($3$), and Categorizations ($2$) and Knowledge Subjects ($6$). The dataset also includes any derived data and figures referenced in this paper. The code to produce those is included as well. The dataset is available at \url{https://hf.co/datasets/matybohacek/benchmarking-cultures-25}.

\newcolumntype{L}[1]{>{\RaggedRight\arraybackslash}p{#1}}

\subsection{Models}

\begin{longtable}{L{4cm} L{11cm}}
\toprule
\textbf{Field} & \textbf{Description} \\
\midrule
\endfirsthead
\toprule
\textbf{Field} & \textbf{Description} \\
\midrule
\endhead

model\_id & Unique identifier for the model (slug). \\
\midrule
model\_name & The display name of the model. \\
\midrule
model\_family & The name of the model family, e.g. Gemini or DeepSeek. \\
\midrule
model\_version & The version of the model, e.g. 2.5 or V3.1. \\
\midrule
model\_variant & The variant of the model, e.g. Flash or Terminus. \\
\midrule
model\_subvariant & A subvariant of the model, e.g Lite. \\
\midrule
model\_is\_base & A flag indicating if the model is a base model. \\
\midrule
model\_total\_parameters & The number of total parameters of the model. \\
\midrule
model\_active\_parameters & The number of active parameters of the model. \\
\midrule
model\_href & URL to the model's press release or blog post. \\
\midrule
model\_published\_at & The date the model was released. \\
\midrule
model\_access & The access level of the model. Options: Closed, Open-Weight or Open-Source. \\
\midrule
model\_has\_highlight & A flag indicating if the model has any benchmark highlights in its release announcement. \\
\midrule
organization\_name & The name of the organization releasing this model. \\
\midrule
organization\_sector & The sector of the organization. Options: Industry, Academia or Non-Profit. \\
\midrule
organization\_country & The country of origin of this organization. \\
\midrule
organization\_domain & The domain of influence this organization belongs to. Options: China or West. \\
\bottomrule
\end{longtable}

\subsection{Benchmarks}

\begin{longtable}{L{4cm} L{11cm}}
\toprule
\textbf{Field} & \textbf{Description} \\
\midrule
\endfirsthead
\toprule
\textbf{Field} & \textbf{Description} \\
\midrule
\endhead

benchmark\_id & Unique identifier for the benchmark (slug). \\
\midrule
benchmark\_name & The display name of the benchmark. \\
\midrule
paper\_id & Unique identifier for the paper announcing the benchmark (arXiv ID or custom slug). \\
\midrule
paper\_href & URL to the paper announcing the benchmark. \\
\midrule
paper\_published\_at & The date the paper was published. This was taken as the benchmark release date (version 1 if more than one was provided). \\
\bottomrule
\end{longtable}

\subsection{Highlights}

\begin{longtable}{L{4cm} L{11cm}}
\toprule
\textbf{Field} & \textbf{Description} \\
\midrule
\endfirsthead
\toprule
\textbf{Field} & \textbf{Description} \\
\midrule
\endhead

benchmark\_id & Unique identifier for the benchmark (slug). \\
\midrule
model\_id & Unique identifier for the model (slug). \\
\midrule
prescribed\_competency & The competency that model builders prescribe to this benchmark for that model release. This field remains empty if the model builder didn't assign a competency but highlighted the benchmark anyway. \\
\midrule
prescribed\_category & A generalized categorization of the prescribed\_competency. \\
\bottomrule
\end{longtable}

\subsection{Affiliations}

\begin{longtable}{L{4cm} L{11cm}}
\toprule
\textbf{Field} & \textbf{Description} \\
\midrule
\endfirsthead
\toprule
\textbf{Field} & \textbf{Description} \\
\midrule
\endhead

paper\_id & Unique identifier for the release paper (arXiv ID or custom slug). \\
\midrule
author\_name & The name of an author for this paper. \\
\midrule
organization\_name & The name of the organization affiliated with the author. \\
\midrule
organization\_sector & The sector of the organization. Options: Industry, Academia or Non-Profit. \\
\midrule
organization\_country & The country of origin of this organization. \\
\midrule
organization\_domain & The domain of influence this organization belongs to. Options: China or West. \\
\bottomrule
\end{longtable}

\newpage
\subsection{Categories}

\begin{longtable}{L{4cm} L{11cm}}
\toprule
\textbf{Field} & \textbf{Description} \\
\midrule
\endfirsthead
\toprule
\textbf{Field} & \textbf{Description} \\
\midrule
\endhead

benchmark\_category & Granular functional classification. Options: Audio-visual pattern recognition, Audio-visual understanding, Coding, Commonsense, Embodied spatial understanding, Factuality, Foundational skills, Generic, Health, Instruction following, Instruction retention, Long-context, Math, Multilingual performance, Multimodal generation, Reasoning and knowledge, Rule adherence, Semantic search, Strategic planning, Tool orchestration, Translation or Writing style. \\
\midrule
benchmark\_meta\_category & High-level classification of the benchmark. Options: Agentic task execution, Formalized comprehension \& reasoning, Information retrieval, Multilingual capabilities, Multimodal processing, Preference-Alignment, Self-contained foundational capabilities, Unstructured comprehension \& reasoning. \\
\midrule
benchmark\_category\_definition & A description of the meaning for the category. \\
\bottomrule
\end{longtable}

\subsection{Categorizations}

\begin{longtable}{L{4cm} L{11cm}}
\toprule
\textbf{Field} & \textbf{Description} \\
\midrule
\endfirsthead
\toprule
\textbf{Field} & \textbf{Description} \\
\midrule
\endhead

benchmark\_id & Unique identifier for the benchmark (slug). \\
\midrule
benchmark\_category & Granular functional classification. Options: Audio-visual pattern recognition, Audio-visual understanding, Coding, Commonsense, Embodied spatial understanding, Factuality, Foundational skills, Generic, Health, Instruction following, Instruction retention, Long-context, Math, Multilingual performance, Multimodal generation, Reasoning and knowledge, Rule adherence, Semantic search, Strategic planning, Tool orchestration, Translation or Writing style. \\
\bottomrule
\end{longtable}

\subsection{Knowledge Subjects}

\begin{longtable}{L{4cm} L{11cm}}
\toprule
\textbf{Field} & \textbf{Description} \\
\midrule
\endfirsthead
\toprule
\textbf{Field} & \textbf{Description} \\
\midrule
\endhead

benchmark\_id & Unique identifier for the benchmark (slug). \\
\midrule
subject & The subject as it was named in the benchmark data set. \\
\midrule
field & A mapping of the subject to a field. Options: Art \& Design, Business, Health \& Medicine, Humanities \& Social Sciences, Law, Science, Tech \& Engineering or nil. \\
\midrule
science\_discipline & A mapping of the science field to a concrete discipline. \\
\midrule
n & The number of questions related to this subject in the benchmark. \\
\midrule
p & The percentage of questions related to this subject in the benchmark. \\
\bottomrule
\end{longtable}

\newpage

\section{Unified Benchmark Taxonomy}
\label{app:unified_taxonomy}

\newcolumntype{L}[1]{>{\RaggedRight\arraybackslash}p{#1}}

\begin{longtable}{L{3.5cm} L{3.5cm} L{8cm}}
\toprule
\textbf{Meta-Category} & \textbf{Category} & \textbf{Definition} \\
\midrule
\endfirsthead

\multirow{4}{=}{\textbf{General knowledge application}} 
    & Reasoning and knowledge & Knowledge retrieval or ``reasoning'' in the sense of solving complex logical problems that ideally are ``non-searchable.'' \\ 
    \cmidrule{2-3}
    & Commonsense & Knowledge and reasoning applied to everyday scenarios rather than specialized domains. \\
\midrule

\multirow{10}{=}{\textbf{Information retrieval}} 
    & Factuality & Testing model knowledge on direct, verifiable facts (e.g., ``What's the capital of France?'') and ability to avoid hallucinations. \\ 
    \cmidrule{2-3}
    & Long-context & Correctly retrieving information from context (e.g., ``Add a paragraph to the poem I asked you to write 10 queries earlier''). \\ 
    \cmidrule{2-3}
    & Semantic search & Tests embedding mechanisms (classifying text based on meaning). Only used when the benchmark explicitly evaluates this. \\
\midrule

\multirow{3}{=}{\textbf{Specialized knowledge application}} 
    & Coding & Code generation, Self-Repair, Code execution. \\ 
    \cmidrule{2-3}
    & Math & Text problems, visual math understanding, result evaluation, process evaluation. \\
\midrule

\multirow{9}{=}{\textbf{Multimodal processing}} 
    & Audio-visual pattern recognition & Simple recognition tasks, such as ``recognize the letters in this image'' or ``count object XYZ.'' \\ 
    \cmidrule{2-3}
    & Audio-visual understanding & Interpretative questions about an image, audio, or video. \\ 
    \cmidrule{2-3}
    & Multimodal generation & Producing audio-visual output (audio, image, video) based on a task. \\ 
    \cmidrule{2-3}
    & Embodied spatial understanding & Three-dimensional orientation and spatial reasoning. \\
\midrule

\multirow{7}{=}{\textbf{Preference-Alignment}} 
    & Generic & Alignment with LLM-judge preferences on an unspecific and broad range of subjects. \\ 
    \cmidrule{2-3}
    & Writing style & Model performance in writing style aligns with LLM-judge preferences. \\ 
    \cmidrule{2-3}
    & Health & Alignment on health-related questions for accuracy and safety (e.g., symptom checking). \\
\bottomrule

\end{longtable}

\textit{Continued on next page...}

\newpage

\begin{longtable}{L{3.5cm} L{3.5cm} L{8cm}}
\toprule
\textbf{Meta Category} & \textbf{Subcategory} & \textbf{Definition} \\
\midrule
\endfirsthead

\multirow{6}{=}{\textbf{Foundational capabilities}} 
    & Instruction following & Explicit evaluation of whether the model correctly follows specific instructions. \\ 
    \cmidrule{2-3}
    & Instruction retention & Ability to maintain state and remember constraints across a multi-turn conversation. \\ 
    \cmidrule{2-3}
    & Base model capabilities & Fundamental aspects of how well the model works as a language model, without targeting a specific downstream application. \\
\midrule

\multirow{7}{=}{\textbf{Agentic task execution}} 
    & Tool orchestration & Checks if models use various tools and their outputs to solve tasks. \\ 
    \cmidrule{2-3}
    & Rule adherence & Checks if the model consistently uses tools in compliance with a rule set. \\ 
    \cmidrule{2-3}
    & Strategic planning & Tasks requiring the identification and execution of intermediate steps to achieve a goal (Chain-of-thought, decomposition). \\
\midrule

\multirow{4}{=}{\textbf{Multilingual capabilities}} 
    & Translation & Translating text or multimodal inputs. \\ 
    \cmidrule{2-3}
    & Multilingual performance & Evaluates model performance across languages in various tasks. \\
\bottomrule
\end{longtable}

\FloatBarrier
\newpage

\section{Full Tables and Figures}
\label{app:full_tables}


\setcounter{table}{9}
\begin{table}[ht]
\centering
\caption{\textbf{Publication Years of Benchmark within Tested Competencies.} Looking at the benchmarks released in 2023, 2024 and 2025 we map the number of benchmarks released per year within a tested competency.}
\label{tab:all_competencies}
\small
\begin{tabular}{llcccccc}
\toprule
& & \multicolumn{2}{c}{\textbf{2023}} & \multicolumn{2}{c}{\textbf{2024}} & \multicolumn{2}{c}{\textbf{2025}} \\
\cmidrule(lr){3-4} \cmidrule(lr){5-6} \cmidrule(lr){7-8}
\textbf{Meta-Category} & \textbf{Category} & $N$ & \% & $N$ & \% & $N$ & \% \\
\midrule
Agentic task execution & Rule adherence & - & - & 1 & 33.3 & 2 & 66.7 \\
Agentic task execution & Strategic planning & - & - & - & - & 12 & 100.0 \\
Agentic task execution & Tool orchestration & - & - & 5 & 38.5 & 7 & 53.8 \\
Foundational capabilities & Base model capabilities & - & - & - & - & 1 & 25.0 \\
Foundational capabilities & Instruction following & 2 & 18.2 & 6 & 54.5 & 3 & 27.3 \\
Foundational capabilities & Instruction retention & - & - & 1 & 50.0 & 1 & 50.0 \\
General knowledge application & Commonsense & - & - & - & - & 1 & 14.3 \\
General knowledge application & Reasoning and knowledge & 10 & 22.2 & 12 & 26.7 & 9 & 20.0 \\
Information retrieval & Factuality & 1 & 9.1 & 5 & 45.5 & 2 & 18.2 \\
Information retrieval & Long-context & - & - & 9 & 75.0 & 3 & 25.0 \\
Information retrieval & Semantic search & 1 & 25.0 & 1 & 25.0 & 1 & 25.0 \\
Multilingual capabilities & Multilingual performance & - & - & 3 & 37.5 & 3 & 37.5 \\
Multilingual capabilities & Translation & 1 & 20.0 & 1 & 20.0 & - & - \\
Multimodal processing & Audio-visual pattern recognition & 3 & 14.3 & 9 & 42.9 & 3 & 14.3 \\
Multimodal processing & Audio-visual understanding & 7 & 15.2 & 22 & 47.8 & 13 & 28.3 \\
Multimodal processing & Embodied spatial understanding & - & - & 1 & 50.0 & 1 & 50.0 \\
Multimodal processing & Multimodal generation & 1 & 10.0 & 2 & 20.0 & 7 & 70.0 \\
Preference-Alignment & Generic & - & - & 2 & 100.0 & - & - \\
Preference-Alignment & Health & - & - & - & - & 2 & 100.0 \\
Preference-Alignment & Writing style & - & - & - & - & 2 & 100.0 \\
Specialized knowledge application & Coding & 3 & 12.0 & 9 & 36.0 & 7 & 28.0 \\
Specialized knowledge application & Math & 11 & 21.6 & 14 & 27.5 & 8 & 15.7 \\
\bottomrule
\end{tabular}
\end{table}

\begin{figure}[h]
    \centering
    \includegraphics[width=1\linewidth]{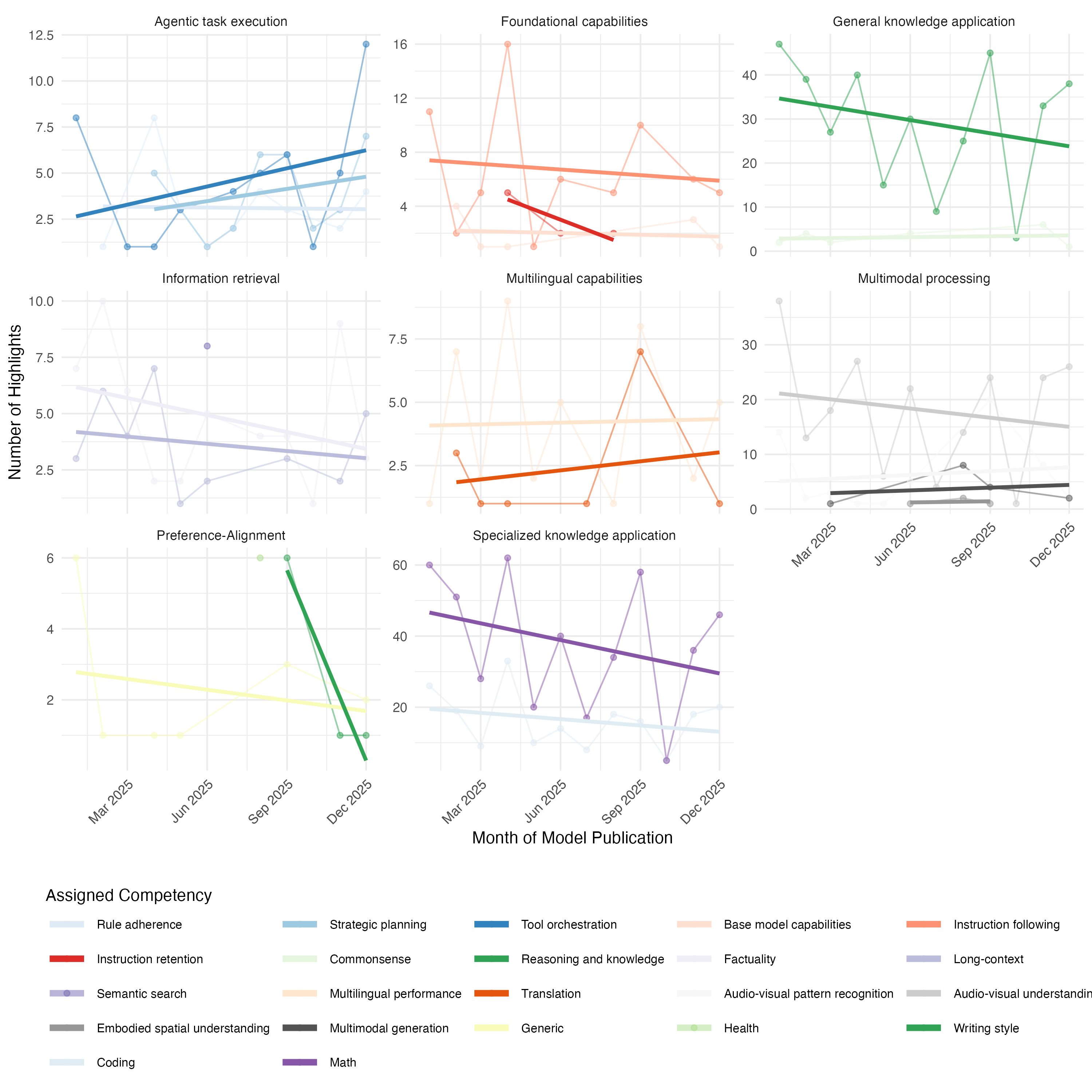}
    \caption{\textbf{Highlights of Competencies by Model Builders.} This graph shows the trend of these selected competencies being highlighted in model releases.}
    \label{fig:highlights}
\end{figure}

\FloatBarrier

\newpage
\FloatBarrier

~

\newpage
\section{\emph{Bench Cultures} Tool Screenshots}
\label{app:bench_cultures_screenshots}

\begin{figure}[h]
    \centering
    \includegraphics[width=0.8\linewidth]{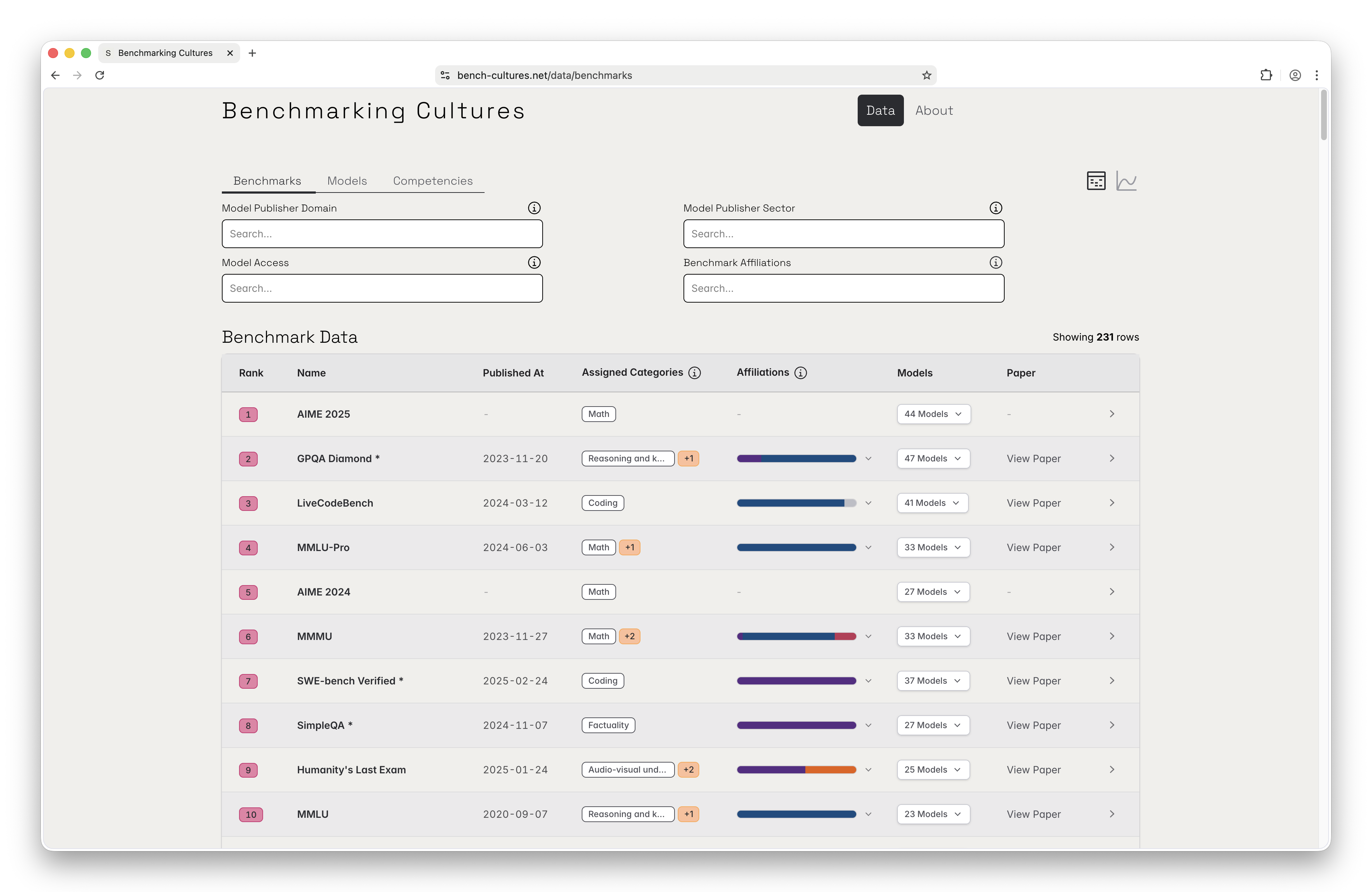}
    \caption{\textbf{Benchmarks View.} Ordered by rank, each benchmark record presents its date of publication, assigned categories and models, affiliation distribution, and a paper link.}
    \label{fig:screen_0a}
\end{figure}

\begin{figure}[h]
    \centering
    \includegraphics[width=0.8\linewidth]{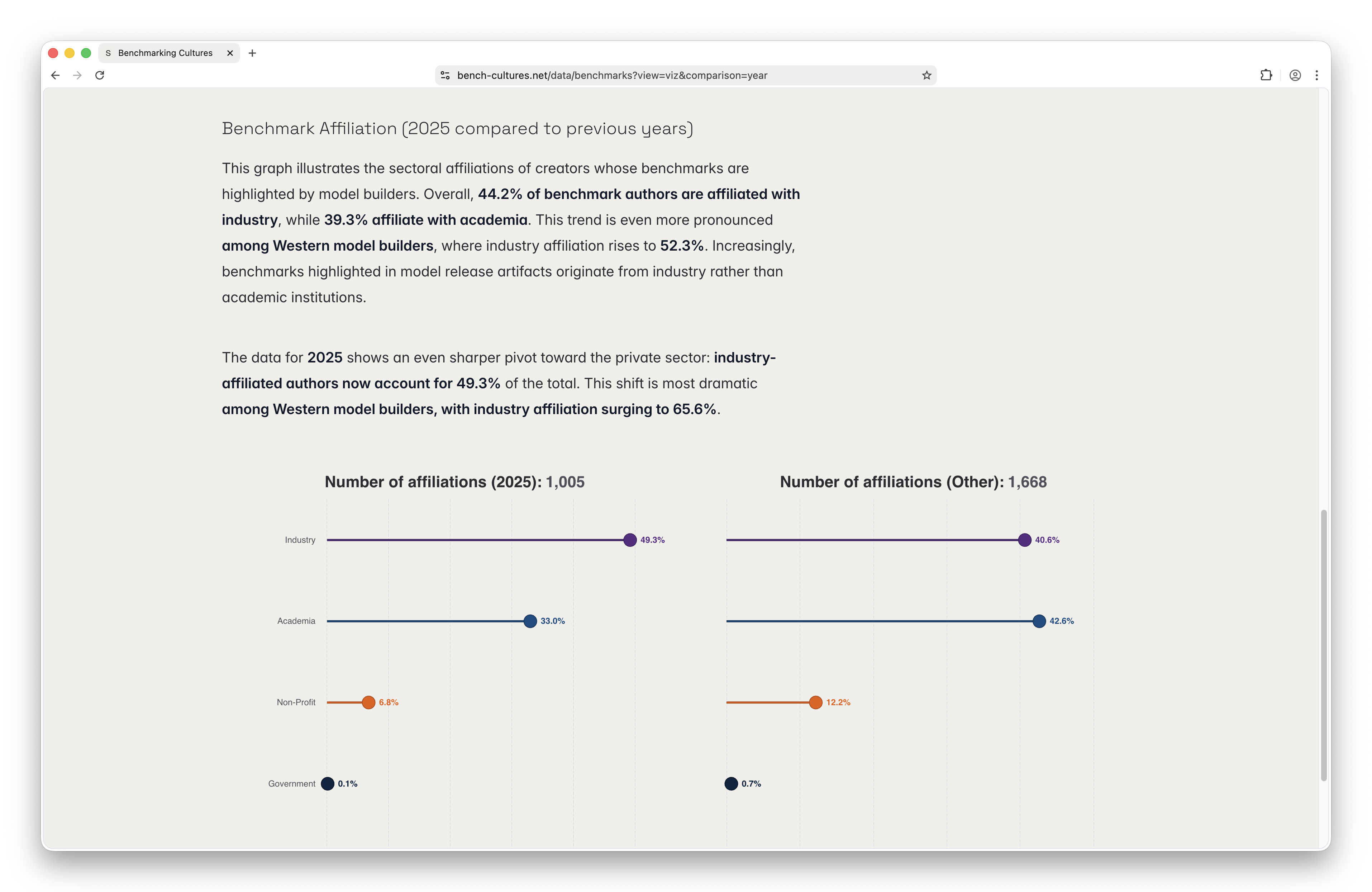}
    \caption{\textbf{Benchmarks Visualization.} Pictured above is a lollipop chart comparison of affiliation of benchmark creators by year, opened from the Benchmarks View.}
    \label{fig:screen_0b}
\end{figure}

\begin{figure}[h]
    \centering
    \includegraphics[width=0.8\linewidth]{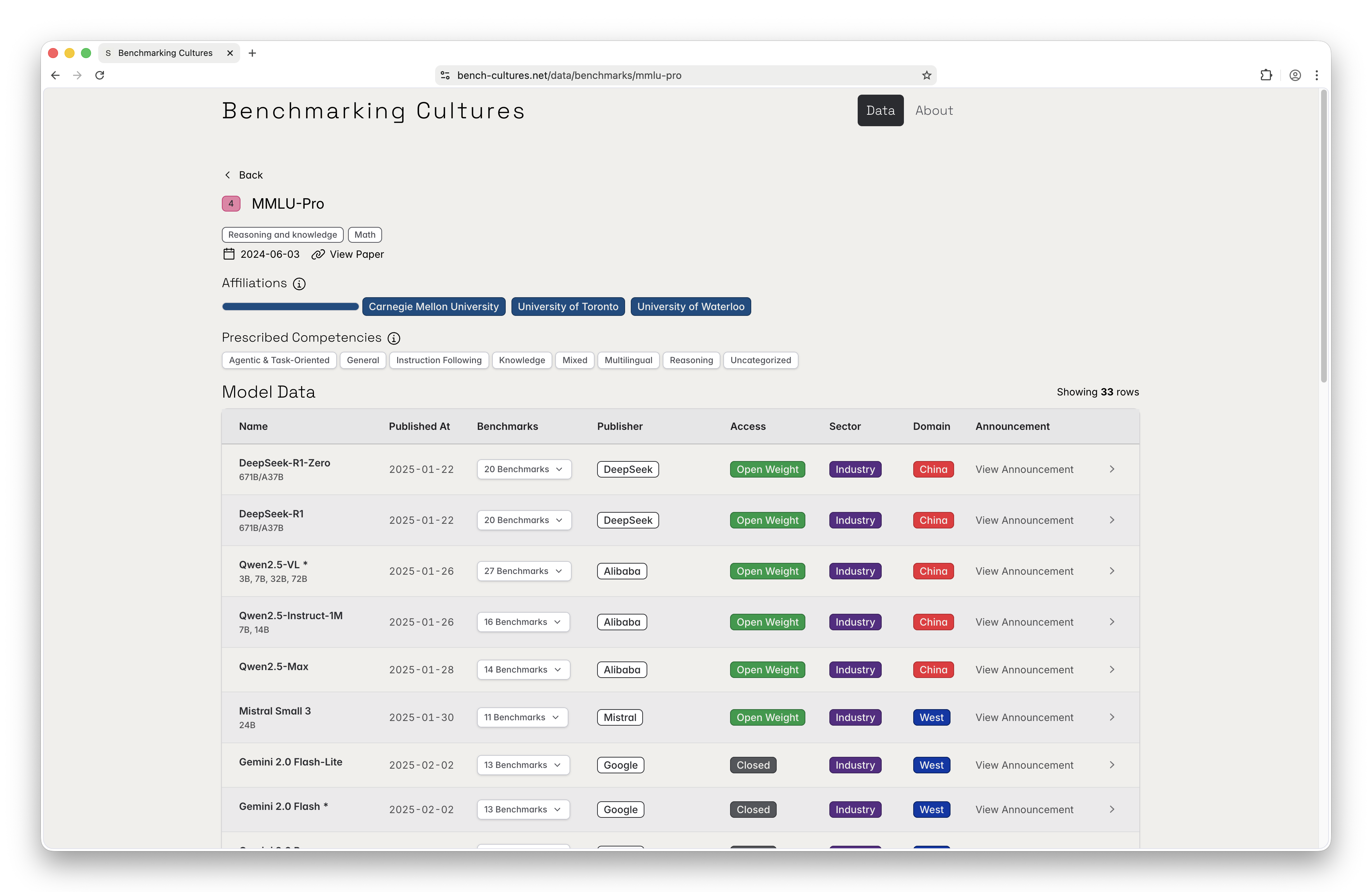}
    \caption{\textbf{Models View.} Pictured above is the models view filtered by MMLU-Pro usage. Each model record presents its date of publication, publisher, access policy, affiliation sector and model parameters if available, domain, and the announcement link.}
    \label{fig:screen_1a}
\end{figure}

\begin{figure}[h]
    \centering
    \includegraphics[width=0.8\linewidth]{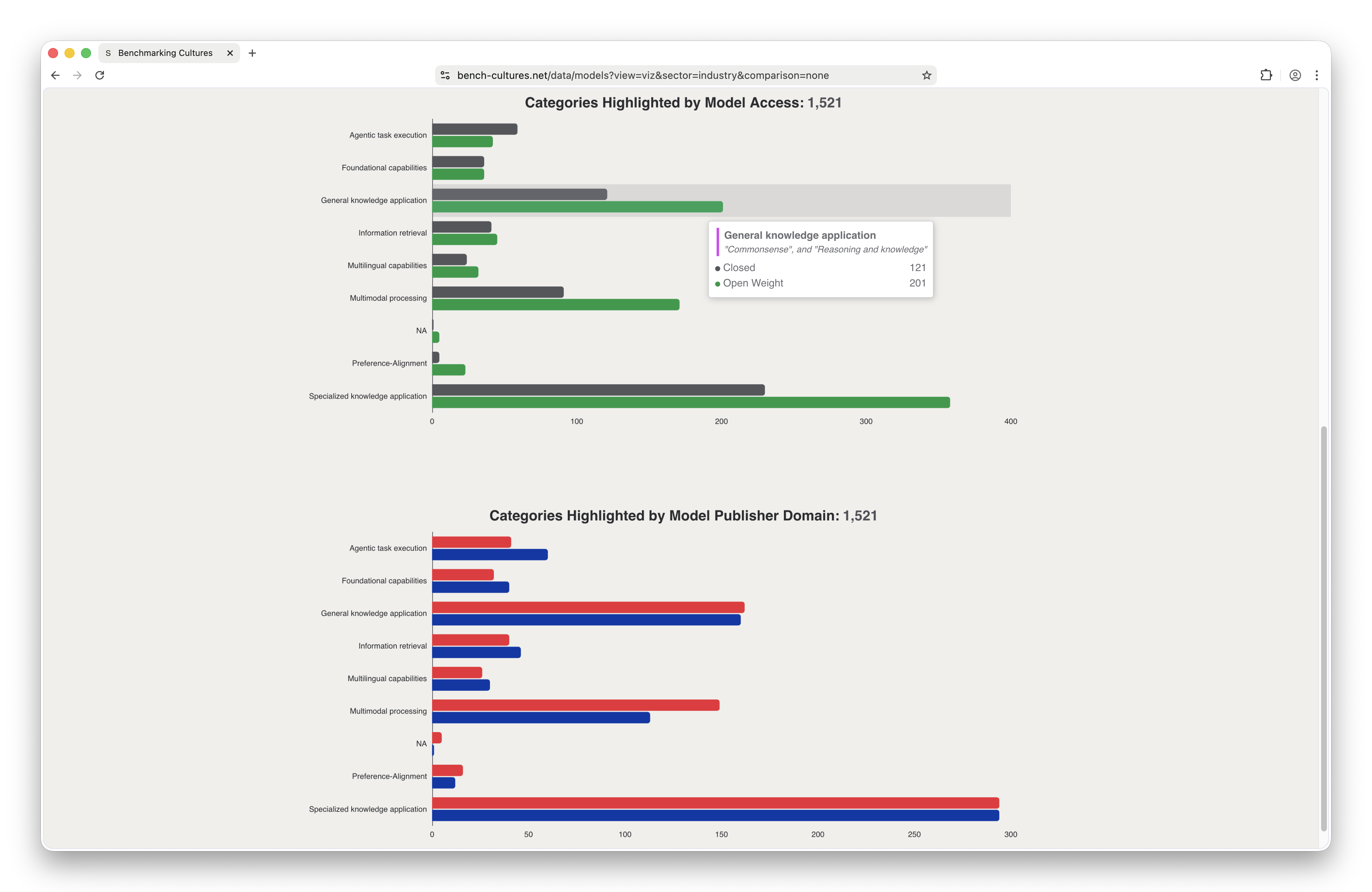}
    \caption{\textbf{Models Visualization.} Pictured above is a grouped bar chart of model access and publisher domain statistics filtered by model publisher sector (Industry), opened from the Models View.}
    \label{fig:screen_1b}
\end{figure}

\begin{figure}[h]
    \centering
    \includegraphics[width=0.8\linewidth]{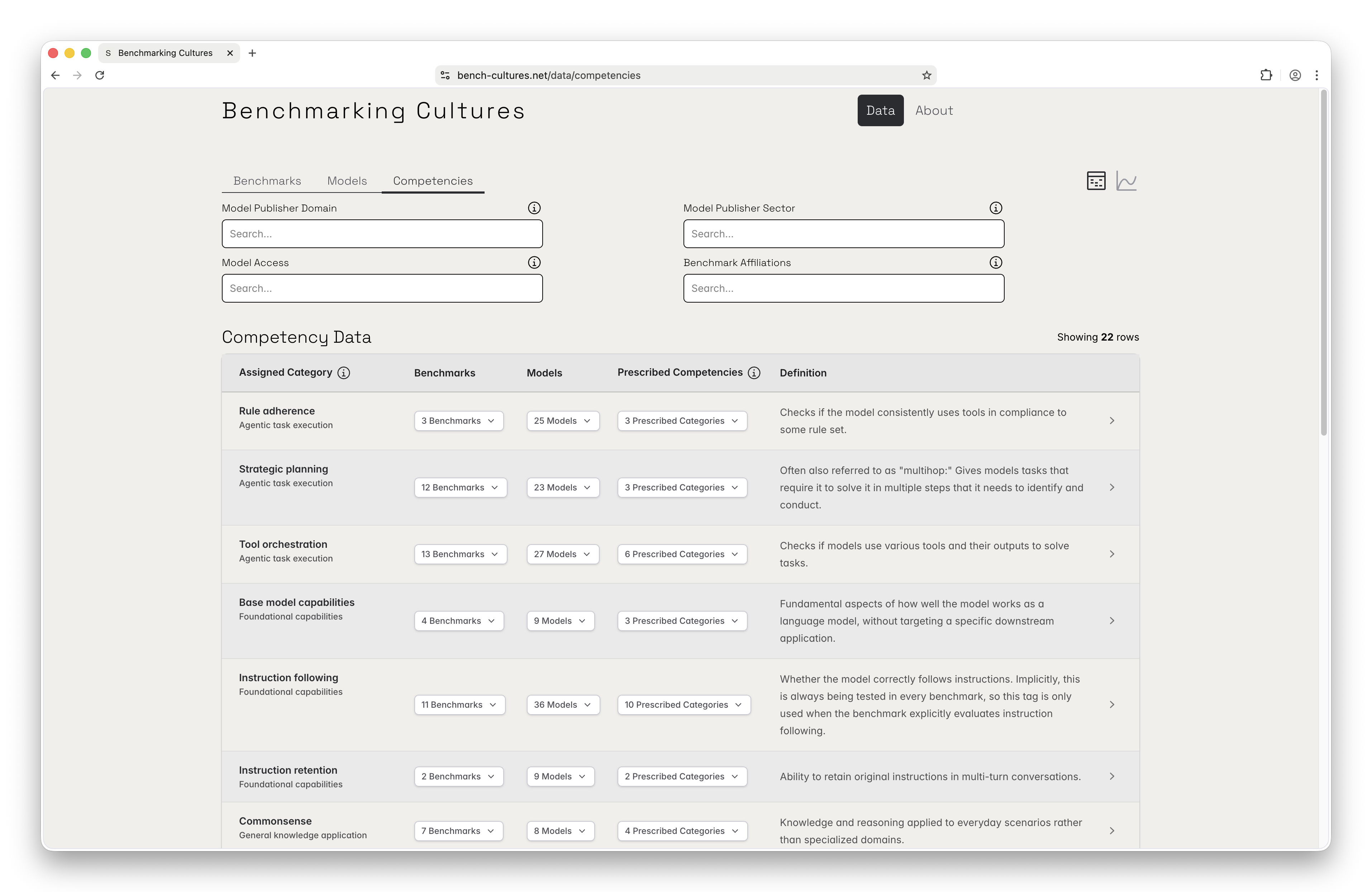}
    \caption{\textbf{Competencies View.} The list contains all tested competencies within our custom taxonomy. Each taxonomy record presents the connected benchmarks, models, and prescribed categories, as well as the definition.}
    \label{fig:screen_2a}
\end{figure}

\begin{figure}[h]
    \centering
    \includegraphics[width=0.8\linewidth]{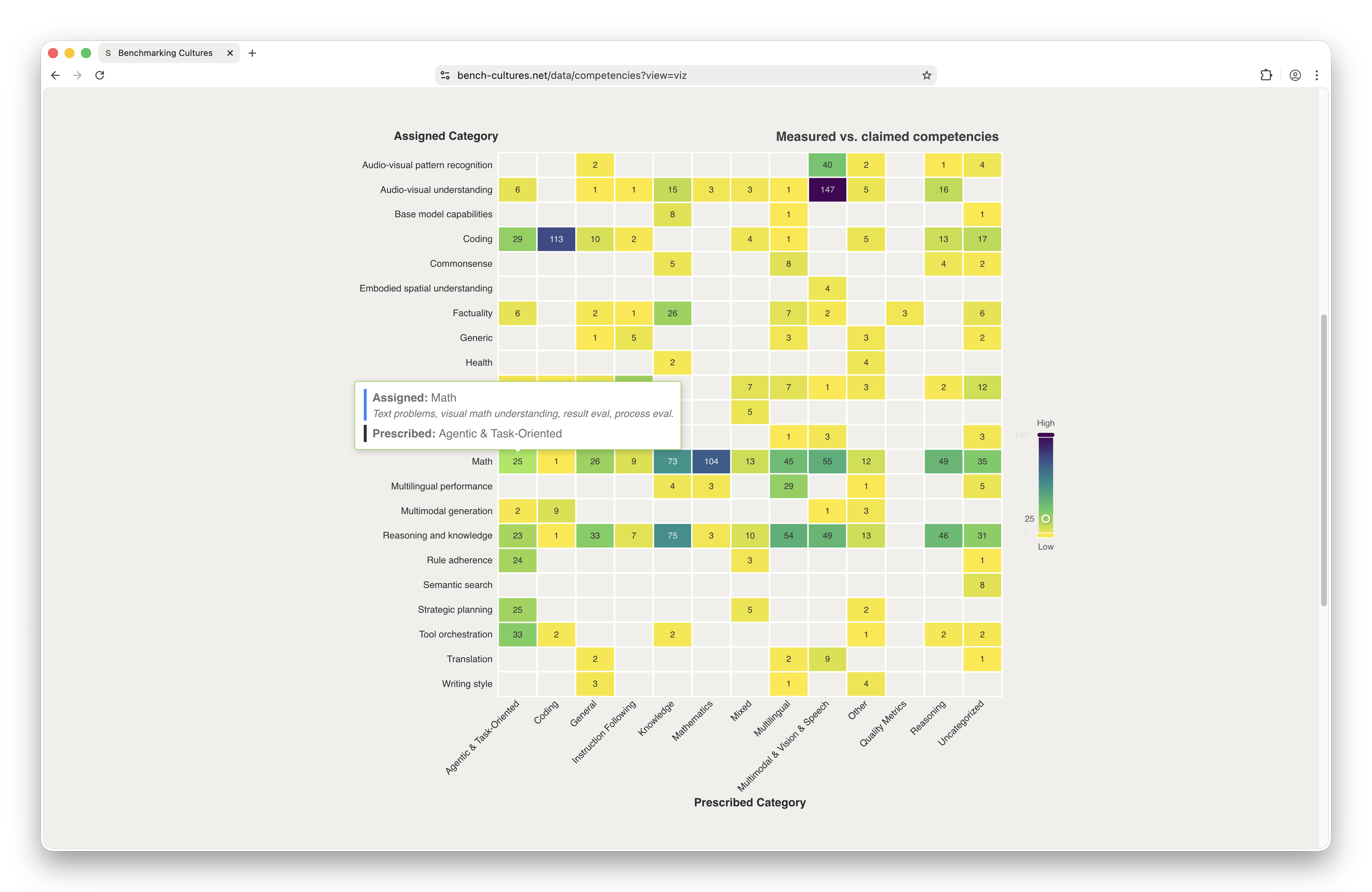}
    \caption{\textbf{Competencies Visualization.} Pictured above is a heatmap chart comparing the competencies that benchmarks are measuring vs. the competencies that model builders prescribe to them, opened from the Competencies View.}
    \label{fig:screen_2b}
\end{figure}

\end{document}